\documentclass[10pt,twocolumn,letterpaper]{article}

\usepackage{cvpr}              
\usepackage{multirow}
\usepackage{makecell}
\usepackage{color}
\usepackage{colortbl}
\usepackage{booktabs}
\usepackage{subcaption}
\usepackage[dvipsnames]{xcolor}

\definecolor{cvprblue}{rgb}{0.21,0.49,0.74}
\usepackage[pagebackref,breaklinks,colorlinks,citecolor=cvprblue]{hyperref}

\def\paperID{005} 
\def\confName{CVPR}
\def\confYear{2024}

\title{XIMAGENET-12: An Explainable Visual Benchmark Dataset for Model Robustness Evaluation}

\author{Qiang Li$^{1,*}$\\
RWTH Aachen, Germany\\
{\tt\small qiang.li@rwth-aachen.de}
\and
Dan Zhang$^{1}$\\
SensLab Technology, China\\
{\tt\small dannie2023.zhang@gmail.com}
\and
Shengzhao Lei $^{2}$\\
EPFL, Switzerland\\
{\tt\small shengzhaolei@gmail.com}
\and
Xun Zhao$^{2}$\\
University of Amsterdam, Netherlands\\
{\tt\small x.zhao@uva.nl}
\and
Weiwei Li$^{3}$\\
Shanghai Business School, China\\
{\tt\small 23154040112@stu.sbs.edu.cn}
\and
Porawit Kamnoedboon$^{3}$\\
University of Zurich, Switzerland\\
{\tt\small porawit.kamnoedboon@uzh.ch}
\and
Junhao Dong $^{3}$\\
Nanyang Technological University, Singapore\\
{\tt\small junhao003@ntu.edu.sg}
\and
Shuyan Li $^{*}$\\
University of Cambridge, UK\\
{\tt\small s12141@cam.ac.uk }
}

\begin{document}
\maketitle
\begin{abstract}
Despite the promising performance of existing visual models on public benchmarks, the critical assessment of their robustness for real-world applications remains an ongoing challenge. To bridge this gap, we propose an explainable visual dataset, XIMAGENET-12, to evaluate the robustness of visual models. XIMAGENET-12 consists of over 200K images with 15,410 manual semantic annotations. Specifically, we deliberately selected 12 categories from ImageNet, representing objects commonly encountered in practical life. To simulate real-world situations, we incorporated six diverse scenarios, such as overexposure, blurring, and color changes, etc. We further develop a quantitative criterion for robustness assessment, allowing for a nuanced understanding of how visual models perform under varying conditions, notably in relation to the background. We make the XIMAGENET-12 dataset and its corresponding code openly accessible at \url{https://sites.google.com/view/ximagenet-12/home}. 
We expect the introduction of the XIMAGENET-12 dataset will empower researchers to thoroughly evaluate the robustness of their visual models under challenging conditions. 
\end{abstract}
\footnotetext[1]{indicates Co-first Authorship, and $\ast$ indicates Shared Corresponding Author. Acknowledging the great support and sponsorship received from Accenture, V7 Lab and Playground AI. Paper accepted by \href{https://syndata4cv.github.io/}{Synthetic Data for Computer Vision Workshop} @ \href{https://cvpr.thecvf.com/Conferences/2024}{IEEE CVPR 2024}}    
\section{Introduction}
\label{sec:intro}

\begin{figure*}[t]
    \centering
    \begin{minipage}{0.14\textwidth}
        \centering
        \includegraphics[width=\linewidth]{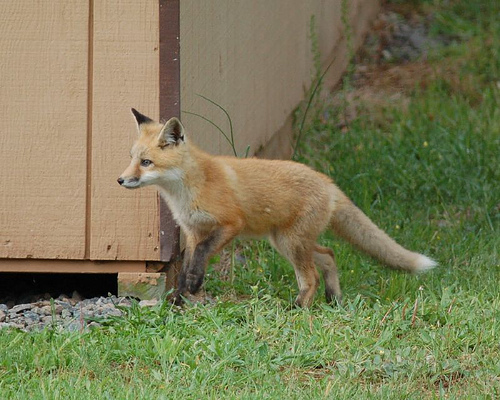}  
        \\\textbf{Original \\image}
    \end{minipage}%
    \hfill
    \begin{minipage}{0.14\textwidth}
        \centering
        \includegraphics[width=\linewidth]{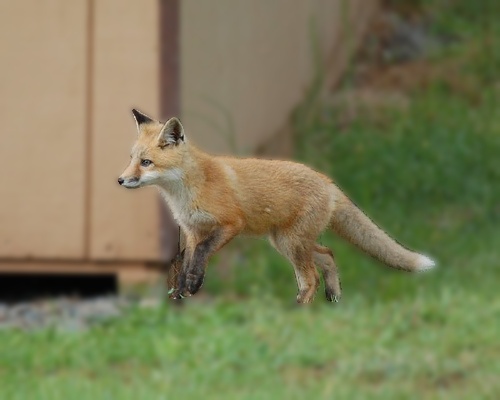}
        \\Blurred background
    \end{minipage}%
    \hfill
    \begin{minipage}{0.14\textwidth}
        \centering
        \includegraphics[width=\linewidth]{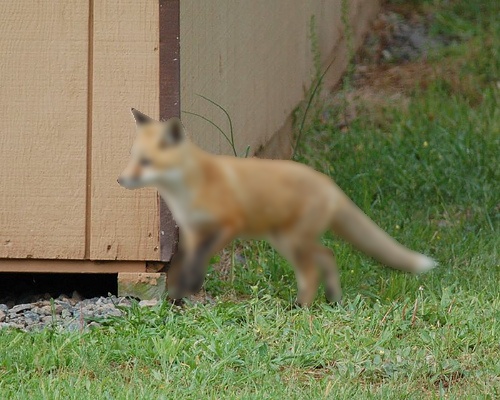}
        \\Blurred foreground
    \end{minipage}%
    \hfill
    \begin{minipage}{0.15\textwidth}
        \centering
        \includegraphics[width=\linewidth]{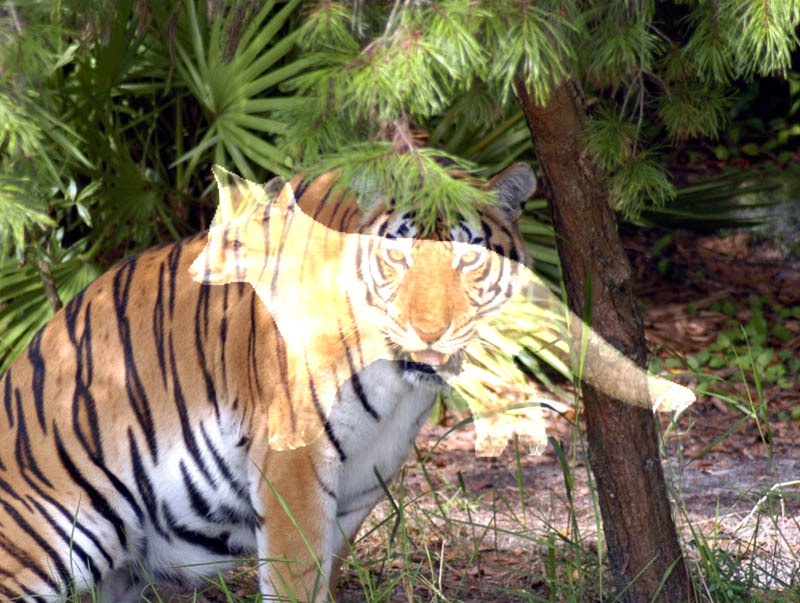}
        \\Random \\generated 
    \end{minipage}%
    \hfill
    \begin{minipage}{0.0925\textwidth}
        \centering
        \includegraphics[width=\linewidth]{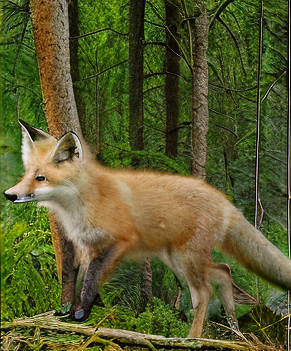}
        \\AI generated
    \end{minipage}%
    \hfill
    \begin{minipage}{0.14\textwidth}
        \centering
        \includegraphics[width=\linewidth]{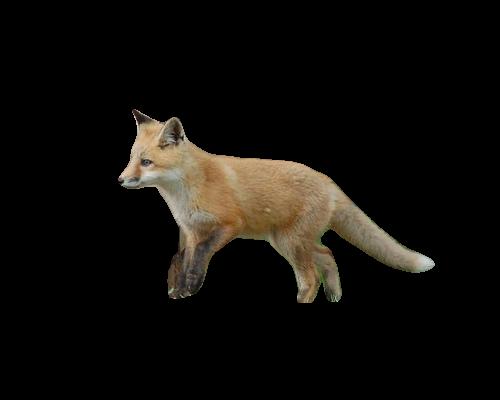}
        \\Segmented image
    \end{minipage}%
    \hfill
    \begin{minipage}{0.14\textwidth}
        \centering
        \includegraphics[width=\linewidth]{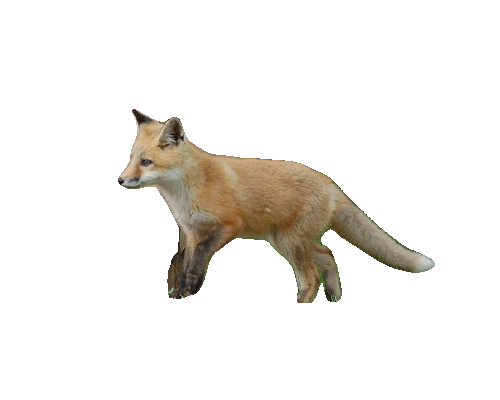}
        \\Transparent image
    \end{minipage}%
    \hfill
    \begin{minipage}{0.14\textwidth}
        \centering
        \includegraphics[width=\linewidth]{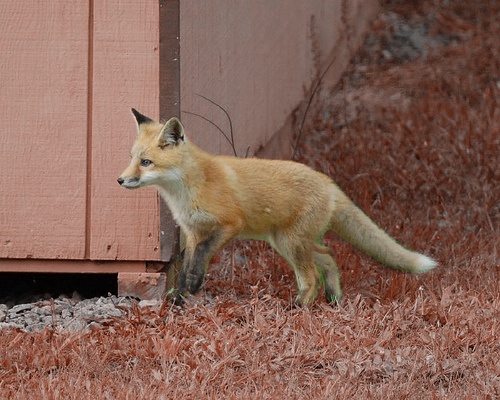}
        \\Color \\R-channel
    \end{minipage}%
    \hfill
    \begin{minipage}{0.14\textwidth}
        \centering
        \includegraphics[width=\linewidth]{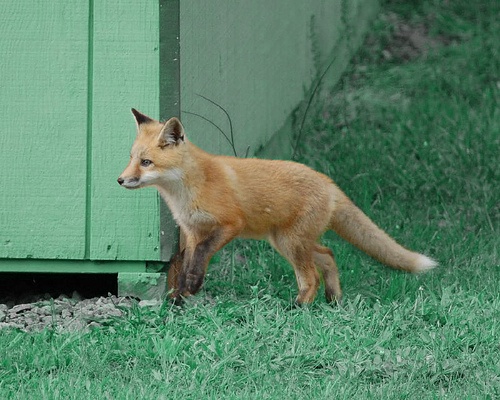}
        \\Color \\G-channel
    \end{minipage}%
    \hfill
    \begin{minipage}{0.14\textwidth}
        \centering
        \includegraphics[width=\linewidth]{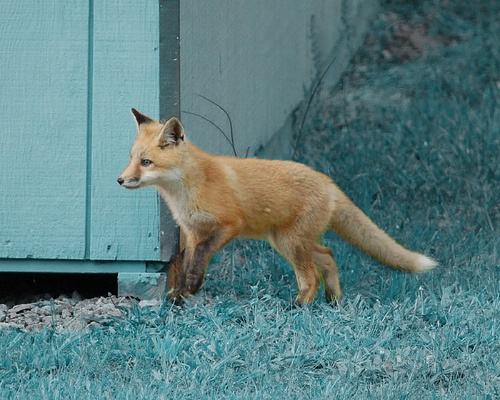}
        \\Color \\B-channel
    \end{minipage}%
    \hfill
    \begin{minipage}{0.14\textwidth}
        \centering
        \includegraphics[width=\linewidth]{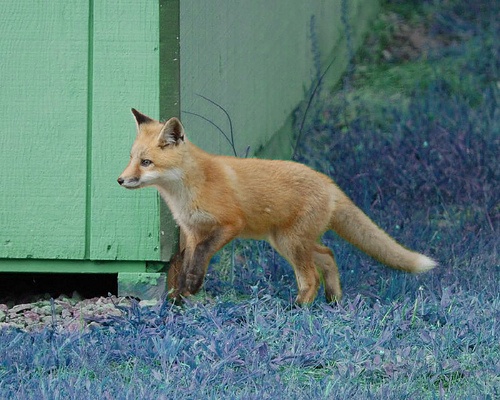}
        \\Color \\rainbow
    \end{minipage}%
    \hfill
    \begin{minipage}{0.14\textwidth}
        \centering
        \includegraphics[width=\linewidth]{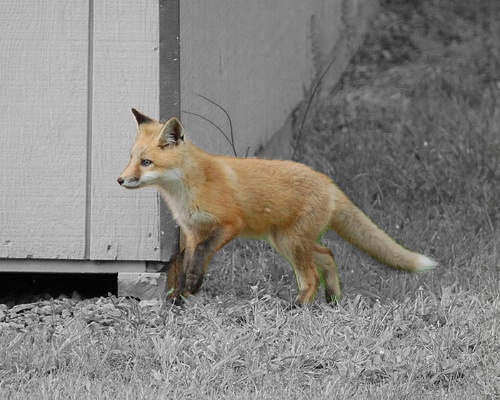}
        \\Color \\grayscale
    \end{minipage}%
    \hfill
    \begin{minipage}{0.14\textwidth}
        \centering
        \includegraphics[width=\linewidth]{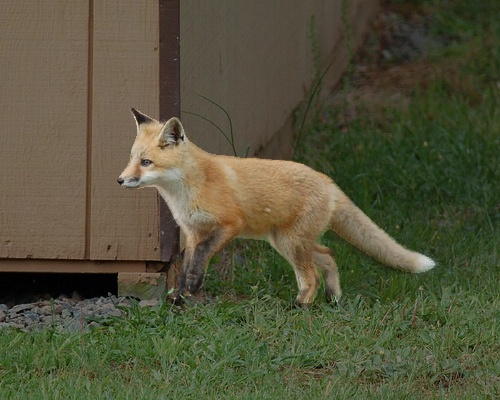}
        \\Bright changing BG
    \end{minipage}%
    \hfill
    \begin{minipage}{0.14\textwidth}
        \centering
        \includegraphics[width=\linewidth]{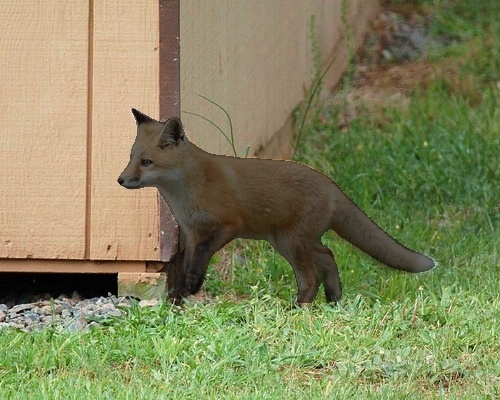}
        \\Bright Changing FG
    \end{minipage}%
    \hfill
    \caption{XIMAGENET-12 sample for 6 scenarios: Blur, Random generated background, AI-generated background, Segmentated, Transparent and Color images. Over 200K Images in total.}
    \label{fig: sample}
\end{figure*}
Visual models have been widely utilized in a variety of real-world applications, including manufacturing, maintenance, etc.~\cite{cvsystem, anomly,zhou2023high,defect,linear}. Despite their commendable performance on standardized benchmark datasets, existing visual models often exhibit noticeable performance degradation in real-world deployments~\cite{cvsystem, zhang2023mm, HGPHH, chen2024taskclip}. Challenges such as variations in lighting, background interference, object displacements and unexpected environmental factors, like noises or artificial camera disturbances, are common issues encountered by visual models in practical scenarios~\cite{linear,anomly,zhang2024unleashing}.


The lack of a readily available and interpretable dataset makes the evaluation of robustness an open challenge. There are a few works attempting to explore how existing visual models are influenced by contextual bias or backgrounds~\cite{DBLP:journals/ijcv/ZhangMLS07,DBLP:conf/kdd/Ribeiro0G16,DBLP:conf/ijcai/ZhuXY17,DBLP:journals/corr/abs-1911-08731,selbst2018intuitive,xiao2020noise}. Among them, the most similar work is ImageNet-9 dataset \cite{xiao2020noise}, which selects nine classes and explores the impact of backgrounds on foreground objects. However, their work did not deeply investigate what factors in the backgrounds really matter for the model behavior, thus leading to less explainability. Besides, their semantic labeling is not precise enough and such rough segmentation of foreground and background leads to some misleading conclusions. 



In this work, we propose an explainable visual benchmark dataset, XIMAGENET-12, to evaluate the robustness of visual models when facing challenging real-world scenarios. XIMAGENET-12
consists of over 200K images with 15,410 manual semantic annotations. Specifically, we deliberately selected 12
categories from ImageNet~\cite{deng2009imagenet}. These images contain objects that are commonly found in daily life, with relatively complicated backgrounds. To simulate real-world situations, we incorporated six diverse scenarios that often occur in real-world applications. As shown in Figure~\ref{fig: sample}, these scenarios cover background \& foreground blurring, color changes to simulate the camera vibrations in industrial production processes, as well as background replacement \& removal and artificially rendered backgrounds for enhanced validation. It is worth noting that our semantic annotations of foreground and background are precise, which allows us to deeply investigate how the visual model is influenced by the backgrounds. We further develop a quantitative criterion for robustness assessment, allowing for a comparative evaluation of visual model robustness. We show that the robustness score of visual backbones calculated on our dataset can provide guidance for practical visual model usage. We summarize our main contributions as follows:

\begin{itemize}
\item We create a dataset, named XIMAGENET-12, consisting of a variety of challenging scenarios. 
\item We develop a quantitative criterion to evaluate the robustness of visual models and show its effectiveness in providing guidance for real-world applications. 
\item We deeply investigate the influence of backgrounds and show some interesting findings based on our well-annotated dataset. 

\end{itemize}

\section{Related Work}
\label{sec:related}

In this section, we discuss previous works that investigate models' performances dependence with contextual bias and backgrounds. Previous research has studied the overarching phenomenon of contextual bias~\cite{torralba2011unbiased,DBLP:conf/eccv/KhoslaZMET12,DBLP:conf/cvpr/ShettySF19}, proposing methods to mitigate its impact. For example, Khosla \itshape et al. \upshape proposed a discriminative framework that directly exploited dataset bias during training~\cite{DBLP:conf/eccv/KhoslaZMET12}. Torralba \itshape et al.\upshape compared multiple popular datasets by using a variety of evaluation criteria to obtain directions that could improve dataset collection and algorithm evaluation protocols~\cite{torralba2011unbiased}. 

Among them, the works most similar to ours are proposed by Zhu \itshape et al.\upshape~\cite{DBLP:conf/ijcai/ZhuXY17} and Xiao \itshape et al.\upshape~\cite{xiao2020noise}, both of which delved into ImageNet\cite{deng2009imagenet} classification and segmentation, and background exploration. Zhu \itshape et al.\upshape~\cite{DBLP:conf/ijcai/ZhuXY17} trained deep neural networks on the foreground and background
respectively, demonstrating that valuable visual hints can be learned separately and then combined to achieve higher
performance. As Zhu \itshape et al.\upshape~\cite{DBLP:conf/ijcai/ZhuXY17} did not conduct the evaluation of recent visual models, Xiao \itshape et al. \upshape~\cite{xiao2020noise} only plainly investigated the influence of backgrounds with state-of-the-art (SOTA) visual models: Noise or Signal. Meanwhile, they proposed a synthetic dataset, ImageNet-9 by disentangling foreground and background signals on ImageNet. Compared with ImageNet-9~\cite{xiao2020noise}, our dataset has more precise semantic labels and we demonstrate that poor semantic label quality can yield sub-optimal results through extensive experiments. Furthermore, our dataset encompasses six scenarios simulating challenges commonly encountered in real-world applications.

\section{XIMAGENET-12 Dataset}
\label{sec:formatting}


\begin{figure*}[t]
    \centering
    \begin{minipage}{0.8\textwidth}
        \centering
        \includegraphics[width=1.0\textwidth]{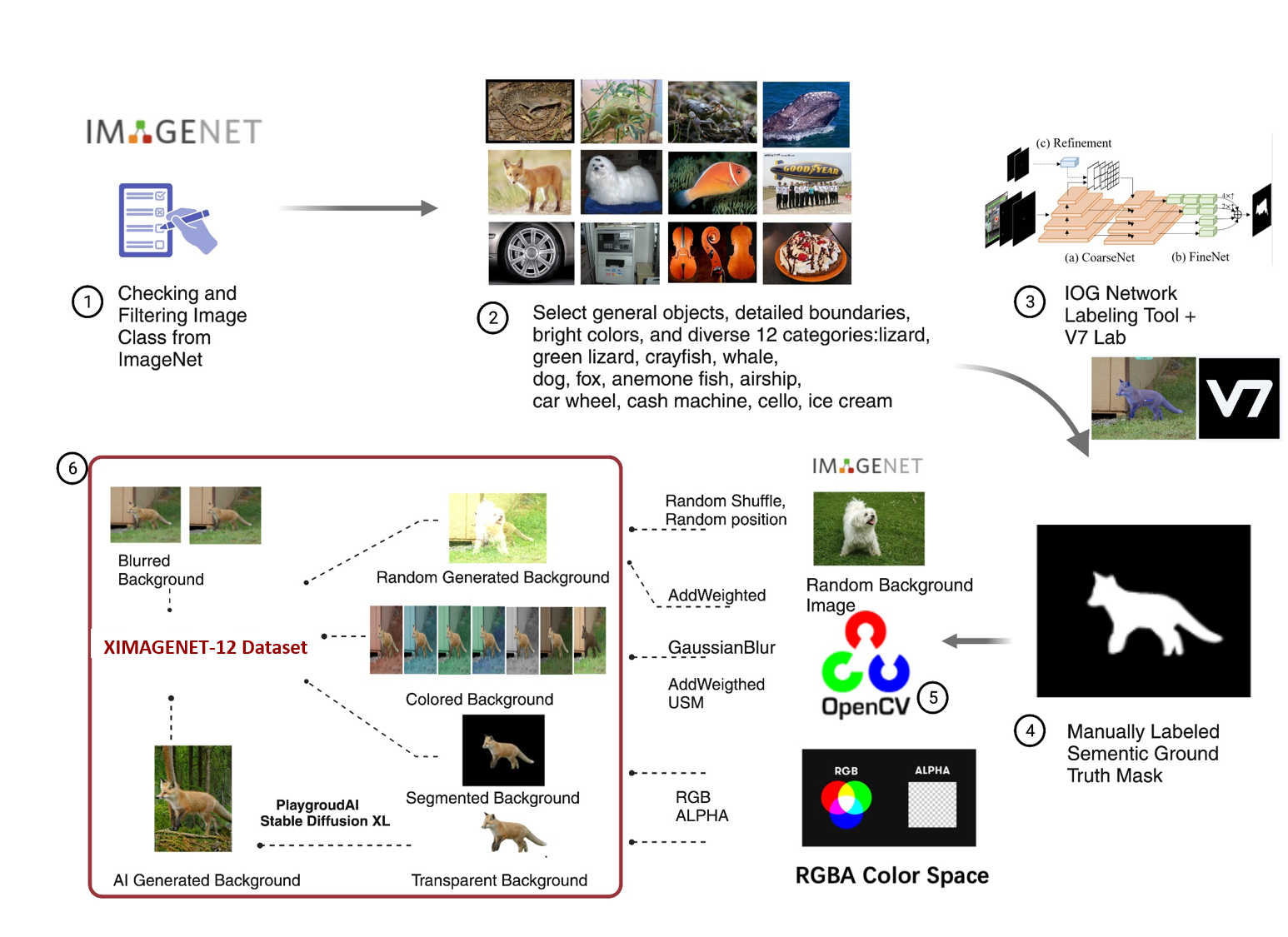} 
    \end{minipage}%
    
    \hfill
    \caption{The flow chart of XIMAGENET-12 generation. }
    \label{fig: overall}
\end{figure*}

\subsection{Dataset Simulation}
\label{ssec:subhead}

The overall dataset generation flow is shown in Figure~\ref{fig: overall}. We select 12 categories of images from the ImageNet~\cite{deng2009imagenet} dataset as the base images. These 12 categories are: \textbf{lizard, green lizard, crayfish, whale, dog, fox, anemone fish, airship, car wheel, cash machine, cello, and ice cream}. These selected images contain objects with complex shapes, detailed edges or boundaries, distinguished colors, diverse resolutions, and complex backgrounds, which can represent situations frequently encountered in daily life. To separate the background and foreground, we employ the IOG network~\cite{zhang2020interactive} for rough segmentation and manually refine the annotations via V7 Lab~\cite{V7Drawin}. Then we synthesize 6 scenarios, including colored images, blurred images, segmented images, transparent images, images with randomly generated backgrounds, and images with AI-generated backgrounds. We detail each scenario as follows.

\textbf{Colored images:} Colored images can simulate lighting changes in the real world. There are 7 different transformations with regards to colored images, including transforming backgrounds to grayscale, single-channel (R, G, B), rainbow, and switching brightness of both backgrounds~(bright changing BG) and foregrounds~(bright changing FG). Specifically, we use the OpenCV function addWeighted to adjust brightness and sharpness via Unsharp Masking (USM). We generate rainbow images by converting the image to HSV color space and changing the hue of the backgrounds. 


\textbf{Blurred images:} Blur often happens when a camera suffers a slight shift, resulting in the degradation of details. We use the OpenCV function GaussianBlur to blur images both in backgrounds~(blurred background) and foregrounds~(blurred foreground). 

\textbf{Segmented images:} We remove the backgrounds of the images and keep the foreground only. Specifically, we keep the RGB channel unchanged for the foreground and set the RGB channel of background as (0, 0, 0).  


\textbf{Transparent images:} We create a new image with RGBA 4 channels, where the background is completely removed. For example, if the pixel at (x, y) in the original image is (r,g,b), we set it as (0, 0, 0, 0) if it belongs to the background, and (r, g, b, 255) vice reverse. 

\textbf{Images with randomly generated backgrounds:} We randomly select an image from ImageNet~\cite{deng2009imagenet} dataset as the background and blend it with the foreground of the original image by using addWeighted blending, resize, random Shuffle \& position functions.  


\textbf{Images with AI-generated backgrounds:} We use the transparent images as inputs for Playground AI~\cite{playgroundAI} with the Stable Diffusion XL model~\cite{podell2023sdxl}. We have tried different text prompts and found some very useful tips: Using keywords such as ‘National Geographic Magazine’ or ‘National Oceanic Magazine’ can increase the authenticity of the generated background; Adding specific and appropriate environmental information to the prompt can make the generated background and objects better integrated. By using a diffusion model and introducing unexpected or extreme background variations, we can assess whether the model is resilient against potential adversarial attacks involving background manipulation.



\begin{figure}[h]
    \centering
    \begin{minipage}{0.5\textwidth}
        \centering
        \includegraphics[width=\linewidth]{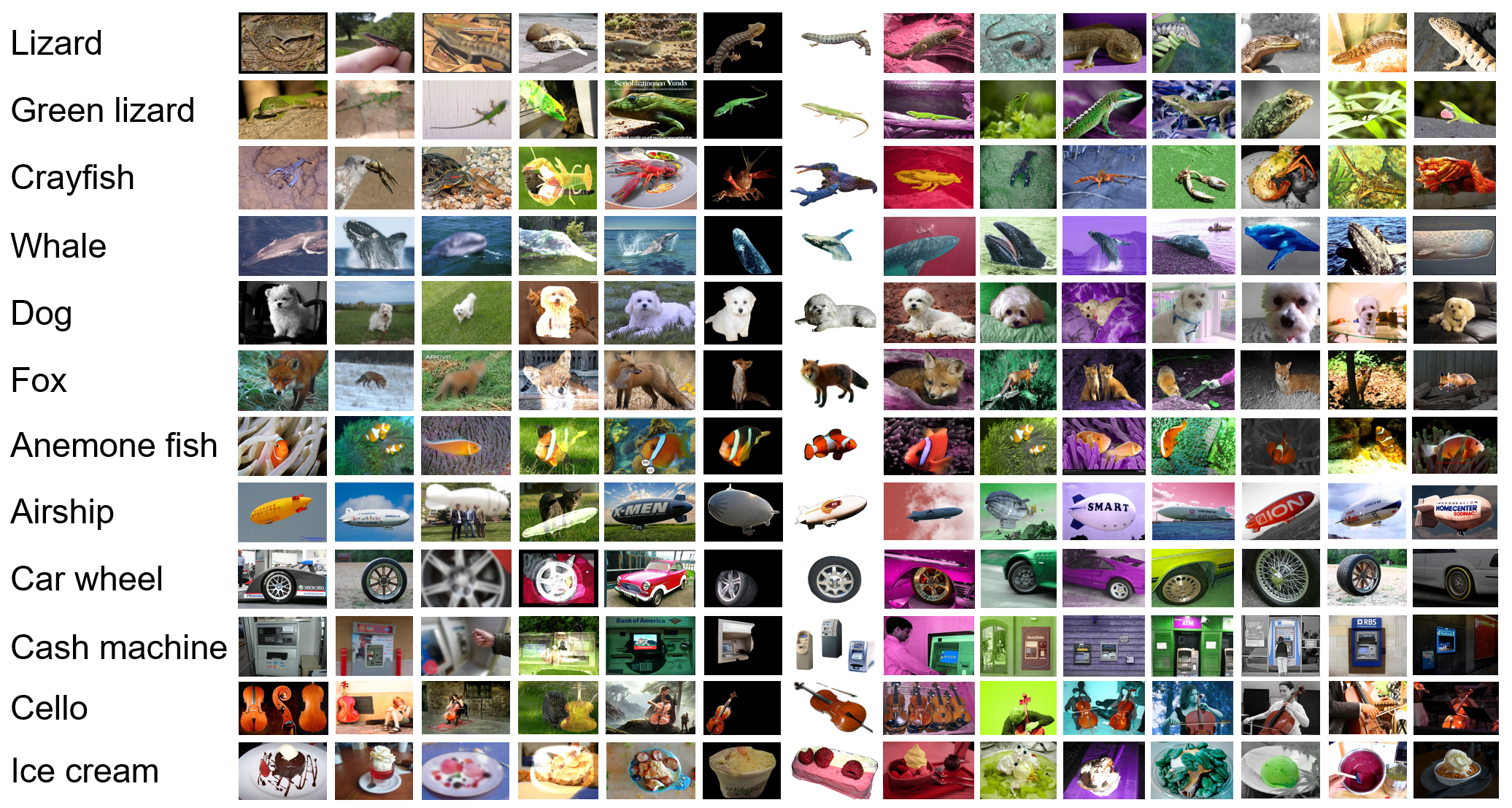}
    \end{minipage}%
    \hfill
    \caption{XIMAGENET-12 samples for each class and scenario.}
    \label{fig:XIMAGE-12 class}
\end{figure}

\subsection{Dataset Properties}
\label{ssec:subhead}
 There are 15,410 original images in XIMAGENET-12, with around 1,300 samples in each category. Each image is simulated to 6 scenarios. As we found that some AI-generated images contained too small or incomplete objects, we filtered those sup-optimal images and obtained 12,248 images for AI-generated scenarios, with approximately 1,000 images for each class. Finally, we have in total 212,747 images contained in XIMAGENET-12. Figure~\ref{fig:XIMAGE-12 class} illustrates the overall look of our dataset across various scenarios, including original images.


\section{Robustness Score}
\label{sec:majhead}

In this section, we introduce our proposed criterion for evaluating the robustness of visual models. We aim to quantitatively assess a model's generalization performance across diverse scenarios. Drawing inspiration from mathematical concepts like variance and covariance, we have devised a robustness score based on the XIMAGENET-12 dataset, as outlined below.

Firstly, we measure the robustness of models in cross scenarios in a variance-like form:
\begin{equation}
  \sigma_{cross}^2 = \frac{\sum_{i=1}^{n}(C(i) - \mu)^2} {n} . 
  \label{eq:important}
\end{equation}

Here, $\mu$ means the best weight accuracy when the model is both trained and tested on the original scenario. $C(i)$ means the model is trained on original images but tested on the $i$-th scenario. $i \in \{0, 1, \dots, n\}$ and $n$ is the number of scenarios that we consider. 

Similarly, we formulate the robustness of models when trained and tested within the same scenario as follows: 

\begin{equation}
  \sigma_{inner}^2 = \frac{\sum_{i=1}^{n}(C'(i) - \mu)^2} {n}.
  \label{eq:also-important}
\end{equation}

Here $C'(i)$ means that the model is both trained and tested on the $i$-th scenario. Considering both above-mentioned cases, we derive the robustness score as follows:

\begin{equation}
  S_{\text{robust}} = 1 - (\sigma_{cross}^2 + \sigma_{inner}^2).
  \label{eq:also-important}
\end{equation}
We consider a larger robustness score as an indicator of the higher robustness of the visual model.


\section{Experiments}
\label{sec:exp}

\subsection{Experimental Settings}
In this section, we evaluated the robustness of commonly used visual models with our proposed XIMAGENET-12 dataset and investigated how visual models perform under various conditions. Specifically, we tested classification models with the following selected visual backbones: ResNet~\cite{he2016deep} series, MobileNet~\cite{sandler2018mobilenetv2}, EfficientNet~\cite{tan2019efficientnet} series, InceptionNet~\cite{InceptionNet}, DenseNet~\cite{huang2017densely}, ViT~\cite{dosovitskiy2020image} and Swin Transformers~\cite{liu2021swin}. We included the following segmentation models with the above-mentioned backbones for further evaluation, including PSPNet \cite{zhao2017pyramid}, FPN \cite{lin2017FPN}, UperNet \cite{upernet}, DeepLabV3 \cite{chen2017rethinking} and DeepLabv3plus R50-D8~\cite{chen2018encoder}.

We conducted our experiments by using TensorFlow~\cite{abadi2016tensorflow}, Keras~\cite{KERAS}, PyTorch~\cite{pytorch}, and MMsegmentation Library~\cite{mmseg2020}. For the inputs of classification models, we cropped the images as $224\times224$. For the inputs of segmentation models, we cropped the images as $256\times256$. We trained these models by using the Adam~\cite{kingma2014adam} optimizer under the learning rate of 0.0001, with the epochs of 200 and the batch size of 16. 

We adopted Top-1 accuracy as the major evaluation metric for classification. We performed Multiple Linear regression~\cite{MLR} to evaluate our hypotheses. We utilized the P-value~\cite{P-VALUE} of 95\% CI as the confidence of verified hypotheses. We employed the Variable that accounts for variations across different models, scenarios, and object classes. Estimate in the Table \ref{tab:CombinedTable} serves as a valuable indicator for assessing accuracy change compared with the reference model. For segmentation models, we used Mean Intersection over Union~(MIoU) and accuracy.



\begin{table*}[t]
\centering
\caption{Comparison of SOTA visual models with diverse scenarios. Here all the evaluation metrics are Top-1 Accuracy. }
\label{tab:Performance}

\resizebox{0.94\textwidth}{!}{%
\begin{tabular}{p{0.2\linewidth}ccccccccccc}
\toprule
\multirow{2}{*}{\textbf{Pretrained Dataset}} & \multirow{2}{*}{\textbf{Model Name}} & \multirow{2}{*}{Parameters (M)} & \multicolumn{8}{c}{\textbf{Test Dataset (Top-1 Acc.)}} &  \\
\cmidrule(lr){4-11}
& & & \textbf{Blur\_bg} & \textbf{Blur\_obj} & \textbf{Color\_g} & \textbf{Color\_b} & \textbf{Color\_grey} & \textbf{Color\_r} & \textbf{Rand\_bg} & \textbf{Seg\_img} \\
\midrule
\multirow{7}{*}{\makecell{ImageNet \cite{deng2009imagenet} \\(Original images) \\ EX1} } & ResNet50 \cite{he2016deep} & 25.60 & 90.97\% & 88.17\% & 84.42\% & 86.98\% & 92.13\% & 89.03\% & 22.41\% & 68.55\% \\
& VGG-16 \cite{simonyan2014very} & 138.4 & 89.92\% & 89.91\% & 78.64\% & 70.46\% & 81.48\% & 80.68\% & 24.58\% & 49.62\% \\
& MobileNetV2 \cite{sandler2018mobilenetv2} & 3.5 & 92.34\% & 88.52\% & 85.73\% & 88.67\% & 88.81\% & 89.33\% & 27.14\% & 66.43\% \\
& EfficientNetB0 \cite{tan2019efficientnet} & 5.3 & 91.44\% & 90.86\% & 78.10\% & 82.45\% & 86.44\% & 83.65\% & 25.29\% & 53.56\% \\
& EfficientNetB3 \cite{tan2019efficientnet} & 12.3 & 86.80\% & 84.53\% & 77.99\% & 81.22\% & 83.00\% & 83.85\% & 22.06\% & 69.67\% \\
& DenseNet121 \cite{huang2017densely}& 8.1 & 93.77\% & 88.92\% & 87.39\% & 87.33\% & 93.23\% & 88.21\% & 26.41\% & 69.67\% \\
& ViT \cite{dosovitskiy2020image} & 86.6 & 88.44\% & 90.77\% & 65.87\% & 62.82\% & 70.69\% & 66.53\% & 17.21\% & 49.01\% \\
& Swin \cite{liu2021swin} & 87.76 & 80.97\% & 81.57\% & 64.59\% & 65.91\% & 69.28\% & 64.41\% & 19.43\% & 44.57\% \\
\midrule
\multirow{6}{*}{\makecell{XImageNet-12 \\ (*Scenarios) \\ EX2}} & ResNet50 \cite{he2016deep}& 25.60 & 83.52\% & 80.24\% & 83.61\% & 84.45\% & 84.71\% & 80.40\% & 53.91\% & 85.76\% \\
& VGG-16 \cite{simonyan2014very}& 138.4 & 74.85\% & 71.54\% & 74.18\% & 76.26\% & 77.58\% & 69.91\% & 70.25\% & 73.27\% \\
& AlexNet \cite{krizhevsky2012imagenet} & 61.1 & 81.60\% & 79.95\% & 81.96\% & 81.89\% & 81.31\% & 78.07\% & 46.29\% & 82.00\% \\
& MobileNetV3 \cite{howard2019searching} & 3.50 & 67.36\% & 67.88\% & 72.04\% & 74.25\% & 69.48\% & 64.79\% & 43.33\% & 78.85\% \\
& DenseNet121 \cite{huang2017densely} & 8,1 & 90.79\% & 86.57\% & 88.92\% & 89.96\% & 90.44\% & 87.37\% & 69.58\% & 91.60\% \\
& ViT \cite{dosovitskiy2020image}& 86.56 & 71.51\% & 70.21\% & 74.77\% & 75.96\% & 75.80\% & 71.14\% & 38.01\% & 78.69\% \\
& Swin \cite{liu2021swin} & 87.76 & 72.81\% & 75.02\% & 81.05\% & 81.96\% & 81.63\% & 76.42\% & 13.23\% & 80.64\% \\

\bottomrule
\end{tabular}
}
\end{table*}

\subsection{Main Results}

\textbf{Comparison of SOTA Visual Models.} Here, we investigate the performance of SOTA models facing diverse scenarios. We study two cases: 1) EX1 Cross Scenario. In this setting, we train the classification model on original images and test them in different scenarios respectively. 2) EX2 Within the same Scenario. In this setting, we both train and test the classification model within the same scenario. We report the classification performance~(Top-1 Accuracy) in Table~\ref{tab:Performance}. In general, different scenarios influence these models to different degrees. Among those scenarios, removing the backgrounds and randomly substituting the background result in most performance drops.


In Table~\ref{tab:Performance} "Rand\_bg" scenario of EX1, all these models show poor performance when trained on original images and tested on images with random backgrounds. This indicates that all these models tend to capture significant information from original backgrounds during training. Those random backgrounds in the test set will heavily interrupt the recognition of visual models. Compared with EX1, the performance drop of EX2 is not so significant. This indicates that when trained on images with random backgrounds, visual models may be aware of the irrationality of the backgrounds and automatically ignore them. 

In Table~\ref{tab:Performance} ``Seg\_img" scenario of EX2, we find that training on the images with removed backgrounds does not lead to a test accuracy drop. This finding contrasts with the assertion by Xiao et al.~\cite{xiao2020noise}, who claimed that removing the background negatively impacts test accuracy. The sub-optimal performance obtained by Xiao et al. is due to the poor annotation quality of ImageNet-9 instead of the missing background~(we will further validate this in Sec~\ref{gen_insti}). Our experiment on XIMAGENET-12 indicates that \itshape \textbf{models trained and tested with well-segmented foregrounds tend to perform well even if the backgrounds are missing}\upshape. 



By observing blurring and color scenarios in Table~\ref{tab:Performance}, CNN-based models generally show better performance with EX1 setting than EX2, while transformer-based models show the opposite results in color scenarios. When observing the Color scenarios (Color\_g to Color\_r) in EX1, CNN-based models show higher robustness (e.g. the drop rate of VGG-16~\cite{simonyan2014very} is 11.28\% ) than transformers~(the drop rate of ViT~\cite{dosovitskiy2020image} is 27.95\%). 

While ViT~\cite{dosovitskiy2020image} and Swin-Transformers~\cite{liu2021swin} show good performance in most visual tasks~\cite{zendel2013vitro}, their accuracy and robustness are not always as good as CNN-based models when facing challenging scenarios. This motivates us with a hypothesis that \itshape \textbf{a model with higher accuracy is not necessarily more stable}\upshape . We argue that \itshape \textbf{more robust models tend to rely less on backgrounds}\upshape. To validate our hypothesis, we provide a deeper investigation of the robustness of existing visual models by using our robustness score and statistical analysis in the following parts.

\begin{table*}[t]\Large
\centering
\caption{Variance of Model Accuracy Performance and Robustness Scores.}
\label{sample-table-scores}
\resizebox{\textwidth}{!}{%

\begin{tabular}{ccccccccccccc}
    \hline
        \multirow{2}{*}{\centering \textbf{Model Acc.  Drop Volatility}} & \multicolumn{8}{c}{Scenarios}& \multirow{2}{*}{\centering Variance} & \multirow{2}{*} {\centering \makecell[c]{Robustness Score \\ (Our* 0 - 1)}  } & \multirow{2}{*}{\centering \makecell[c]{Offical Top-1 Acc. \\ (On ImageNet~\cite{deng2009imagenet})} }  & \multirow{2}{*}{\centering \makecell[c]{Offical Top-1 Acc. \\ (On Cifar10~\cite{cifar10})} } \\
        \cmidrule(lr){2-9}
        ~ & Blur\_background & Blur\_object & Image\_g & Image\_b & Image\_grey & Image\_r & Random\_background & Segmented\_image & ~ & ~ & ~ \\ \hline
        \\
        ResNet50 \cite{he2016deep}: external & 0,18\% & 0,50\% & 1,17\% & 0,68\% & 0,10\% & 0,38\% & 53,04\% & 7,12\% & 0,0902 & \multirow{2}{*}{0,8985} & \multirow{2}{*}{74,90\%}  & \multirow{2}{*}{93,03\%} \\ 
        ResNet50 \cite{he2016deep}: internal & 0,10\% & 0,00\% & 0,11\% & 0,17\% & 0,20\% & 0,00\% & 6,96\% & 0,30\% & 0,0112 & ~ & ~ \\ 
        \\
        DenseNet121 \cite{huang2017densely}:external & 0,13\% & 0,72\% & 1,00\% & 1,01\% & 0,17\% & 0,84\% & 50,39\% & 7,69\% & 0,0885 & \multirow{2}{*}{0,9062} & \multirow{2}{*}{75,00\%} & \multirow{2}{*}{96,54\%}  \\ 
        DenseNet121 \cite{huang2017densely}:internal & 0,21\% & 0,00\% & 0,07\% & 0,14\% & 0,18\% & 0,01\% & 2,77\% & 0,29\% & 0,0052 & ~ & ~ \\ 
        \\
        VGG-16 \cite{simonyan2014very}:external & 0,15\% & 0,15\% & 2,30\% & 5,45\% & 1,52\% & 1,72\% & 47,93\% & 19,53\% & 0,1125 & \multirow{2}{*}{0,8845} & \multirow{2}{*}{71,30\%} & \multirow{2}{*}{93,43\%} \\ 
        VGG-16 \cite{simonyan2014very}:internal & 0,33\% & 0,06\% & 0,26\% & 0,51\% & 0,72\% & 0,01\% & 0,01\% & 0,17\% & 0,0029 & ~ & ~ \\ 
        \\
        ViT \cite{dosovitskiy2020image}:external & 0,25\% & 0,07\% & 7,60\% & 9,38\% & 5,18\% & 7,24\% & 58,11\% & 19,74\% & 0,1536 &\multirow{2}{*}{0,8196} & \multirow{2}{*}{81,07\%} & \multirow{2}{*}{98,20\%} \\ 
        ViT \cite{dosovitskiy2020image}:internal & 0,51\% & 0,72\% & 0,15\% & 0,07\% & 0,08\% & 0,57\% & 16,54\% & 0,00\% & 0,0266 \\ 
        \\
        \\
        Swin \cite{liu2021swin}:external & 0,96\% & 0,84\% & 6,84\% & 6,17\% & 4,61\% & 6,94\% & 50,87\% & 21,33\% & 0,1408 &\multirow{2}{*}{0,6305} & \multirow{2}{*}{83.58\%} & \multirow{2}{*}{97,95\%} \\ 
        Swin \cite{liu2021swin}:internal & 0,02\% & 56,28\% & 0,48\% & 0,61\% & 0,56\% & 0,05\% & 37,10\% & 65,03\% & 0,2287 \\ 
        \\
        \hline
\end{tabular}}
\end{table*}

\begin{table*}
\centering
\caption{Performance Evaluation of Multiple Linear Regression. (P value $<$ 0.0001 and **** indicate the result is of high significance. ns note as not significant). }
\resizebox{\linewidth}{!}{%
\begin{tabular}{@{}cccccccc@{}}
\toprule
\multicolumn{4}{c}{\textbf{Classification}} & \multicolumn{4}{c}{\textbf{Segmentation}} \\
\cmidrule(lr){1-4} \cmidrule(lr){5-8}
\textbf{Variable} & \textbf{Estimate} & \textbf{P value} & \textbf{P value summary} & \textbf{Variable} & \textbf{Estimate} & \textbf{P value} & \textbf{P value summary} \\
\midrule
Intercept & 0.8986 & $\textless 0.0001$ & \cellcolor{red!25}**** & Intercept & \textbf{0.6672} & $\textless 0.0001$ & \cellcolor{red!25}**** \\
Model Name[EfficientNetB0 \cite{tan2019efficientnet}] & -0.03444 & 0.0175 & \cellcolor{red!25}* & Model Name[dpt\_vit-b16 \cite{dosovitskiy2020image}] & \textbf{-0.1999} & $\textless 0.0001$ & \cellcolor{red!25}**** \\
Model Name[EfficientNetB3 \cite{tan2019efficientnet}] & -0.04111 & 0.0046 & \cellcolor{red!25}** & Model Name[upernet\_swin \cite{upernet}] & -0.2211 & $\textless 0.0001$ & \cellcolor{red!25}**** \\
Model Name[DenseNet121 \cite{huang2017densely}] & 0.01556 & 0.2821 & ns & Model Name[upernet\_vit-b16 \_ln\_mln \cite{upernet}] & -0.1695 & $\textless 0.0001$ & \cellcolor{red!25}**** \\
Model Name[MobileNetV2 \cite{sandler2018mobilenetv2}] & 0.005556 & 0.7007 & ns & Model Name[pspnet\_r50-d8 \cite{zhao2017pyramid}] & -0.045 & 0.0106 & \cellcolor{red!25}* \\
 &  &  &  & Model Name[fpn\_r50 \cite{lin2017FPN}] & -0.2081 & $\textless 0.0001$ & \cellcolor{red!25}**** \\
 &  &  &  & Model Name[upernet\_r50 \cite{upernet}] & -0.05796 & 0.001 & \cellcolor{red!25}** \\
\addlinespace
Image Scenario[blur\_background] & -0.0425 & 0.0288 & \cellcolor{red!25}* & Image Scenario[blur\_background] & 0.01833 & 0.3576 & ns \\
Image Scenario[blur\_object] & -0.07 & 0.0003 & \cellcolor{red!25}*** & Image Scenario[blur\_object] & -0.1571 & $<0.0001$ & \cellcolor{red!25}**** \\
Image Scenario[image\_g] & -0.1257 & $<0.0001$ & \cellcolor{red!25}**** & Image Scenario[image\_g] & -0.07131 & 0.0004 & \cellcolor{red!25}*** \\
Image Scenario[image\_b] & -0.0985 & $<0.0001$ & \cellcolor{red!25}**** & Image Scenario[image\_b] & -0.03952 & 0.0476 & \cellcolor{red!25}* \\
Image Scenario[image\_grey] & \textbf{-0.06517} & 0.0008 & \cellcolor{red!25}*** & Image Scenario[image\_grey] & -0.01929 & 0.3332 & ns \\
\addlinespace
Image Scenario[image\_r] & -0.087 & $<0.0001$ & \cellcolor{red!25}**** & Image Scenario[image\_r] & -0.07702 & $0.0001$ & \cellcolor{red!25}*** \\
Image Scenario[Random Background with Real Environment] & \textbf{-0.7078} & $<0.0001$ & \cellcolor{red!25}**** & Image Scenario[segmented\_image] & -0.08143 & $<0.0001$ & \cellcolor{red!25}**** \\
Image Scenario[Segmented\_image] & -0.3012 & $<0.0001$ & \cellcolor{red!25}**** & Image Scenario[generated\_background] & -0.1408 & $<0.0001$ & \cellcolor{red!25}**** \\
\addlinespace
Image Class[1] & 0.134 & $<0.0001$ & \cellcolor{red!25}**** & Image Class[1] & 0.07619 & 0.001 & \cellcolor{red!25}*** \\
Image Class[2] & -0.04867 & 0.0301 & \cellcolor{red!25}* & Image Class[2] & -0.06508 & 0.0048 & \cellcolor{red!25}** \\
Image Class[3] & 0.04 & 0.0745 & ns & Image Class[3] & 0.05222 & 0.0234 & \cellcolor{red!25}* \\
Image Class[4] & 0.1004 & $<0.0001$ & \cellcolor{red!25}**** & Image Class[4] & 0.08127 & 0.0004 & \cellcolor{red!25}*** \\
Image Class[5] & 0.1333 & $<0.0001$ & \cellcolor{red!25}**** & Image Class[5] & 0.2713 & $<0.0001$ & \cellcolor{red!25}**** \\
\addlinespace
Image Class[6] & 0.07667 & 0.0007 & \cellcolor{red!25}*** & Image Class[6] & 0.3021 & $<0.0001$ & \cellcolor{red!25}**** \\
Image Class[7] & 0.01044 & 0.6409 & ns & Image Class[7] & 0.1641 & $7.137$ & \cellcolor{red!25}**** \\
Image Class[8] & 0.09067 & $<0.0001$ & \cellcolor{red!25}**** & Image Class[8] & 0.1548 & $6.73$ & \cellcolor{red!25}**** \\
Image Class[9] & 0.09933 & $<0.0001$ & \cellcolor{red!25}**** & Image Class[9] & 0.2216 & $9.635$ & \cellcolor{red!25}**** \\
\addlinespace
Image Class[10] & 0.1651 & $<0.0001$ & \cellcolor{red!25}**** & Image Class[10] & 0.2689 & $11.69$ & \cellcolor{red!25}**** \\
Image Class[11] & -0.02244 & 0.3164 & ns & Image Class[11] & 0.04079 & $1.774$ & ns \\
\bottomrule
\end{tabular}}
\label{tab:CombinedTable}
\end{table*}

\begin{figure}[b]
    \centering
    \begin{minipage}{0.5\textwidth}
        \centering
        \includegraphics[width=1\linewidth]{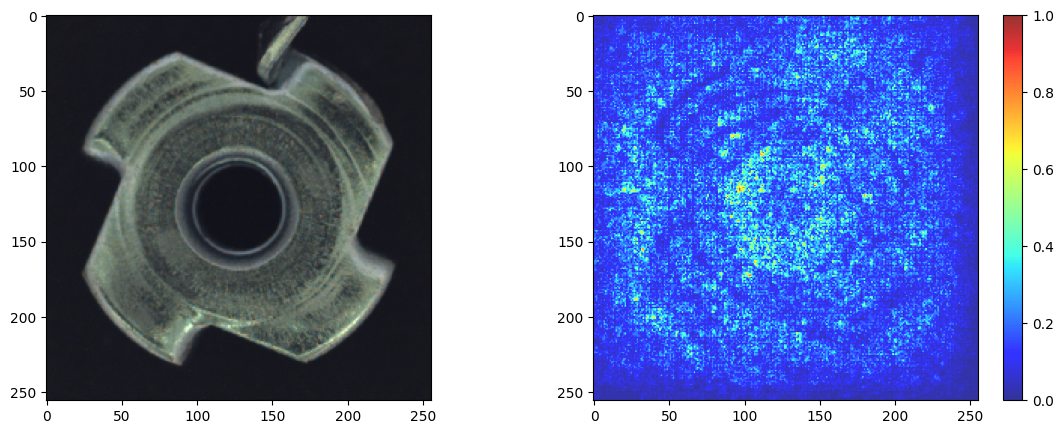}
        \\ ResNet50 \cite{he2016deep} with relatively higher Robustness Score on our Dataset, also perform good in MVTec AD dataset \cite{mvtec}, can detect more details in scratches on metal nut rather than general focus on flaw shape.
        \label{fig:DenseNet}
	\end{minipage}
	\hfill
	\begin{minipage}{0.5\textwidth}
    \centering
    \includegraphics[width=1\linewidth]{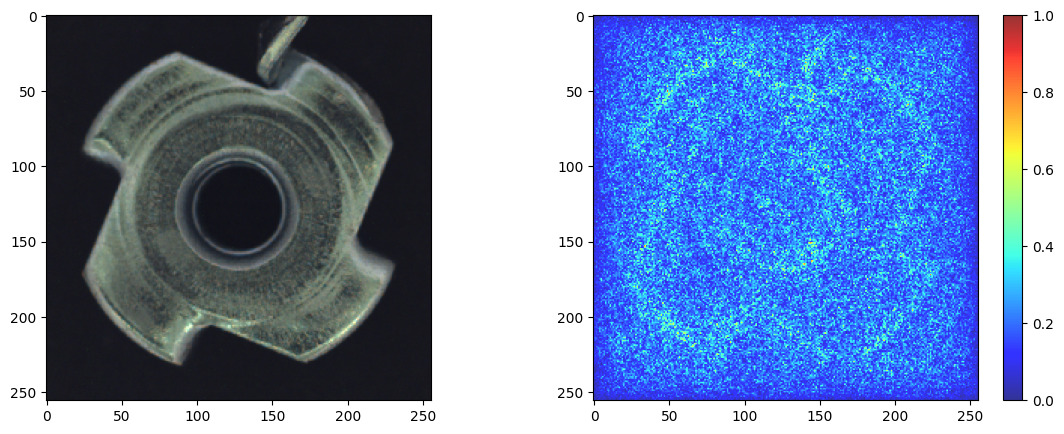}
    \\ VGG-16 \cite{simonyan2014very} with relatively lower Robustness Score on our Dataset, perform poor in MVTec AD dataset \cite{mvtec}.
    \label{fig:ResNet}
	\end{minipage}
	\caption{Saliency Map Analysis on the MVTec AD Dataset \cite{mvtec}.}
   \label{fig:med}  
\end{figure}

\textbf{Evaluation of Robustness.} Using our proposed robustness score, we quantitatively evaluate the robustness of commonly used visual models and report the results in Table~\ref{sample-table-scores}. We find that a model with a higher robustness score is more resistant to the changes in background~(with lower performance variance). Consistent with the observation in the previous part, ViT~\cite{dosovitskiy2020image} and Swin-Transformer~\cite{liu2021swin} have higher variances across diverse scenarios and lower robustness scores. 


Table~\ref{sample-table-scores} also shows the performance on another dataset Cifar10~\cite{cifar10}. In this case, for CNN-based models, a model with a higher robustness score tends to have higher accuracy. For the transformer-based models, Swin-Transformer~\cite{liu2021swin} has both a higher robustness score and better performance. This indicates that the robustness score calculated on XIMAGENET-12 can be an effective performance indicator for other datasets. To show that our robustness evaluation can also provide helpful guidance to real-world applications such as industry, we investigate the performance of ResNet50~\cite{he2016deep} and VGG-16~\cite{simonyan2014very} backbones on an industrial anomaly detection dataset MVTec AD~\cite{mvtec}. We show the Saliency maps of ResNet50~\cite{he2016deep} and VGG-16\cite{simonyan2014very} on MVTec AD~\cite{mvtec} in Figure~\ref{fig:med}. It indicates that the model with a higher robustness backbone tends to focus more on features from the foreground object and less on the background, such as scratches on metal nuts.

\begin{figure*}[t]
    \centering
    \begin{minipage}{0.32\linewidth}
        \centering
        \includegraphics[width=1\linewidth]{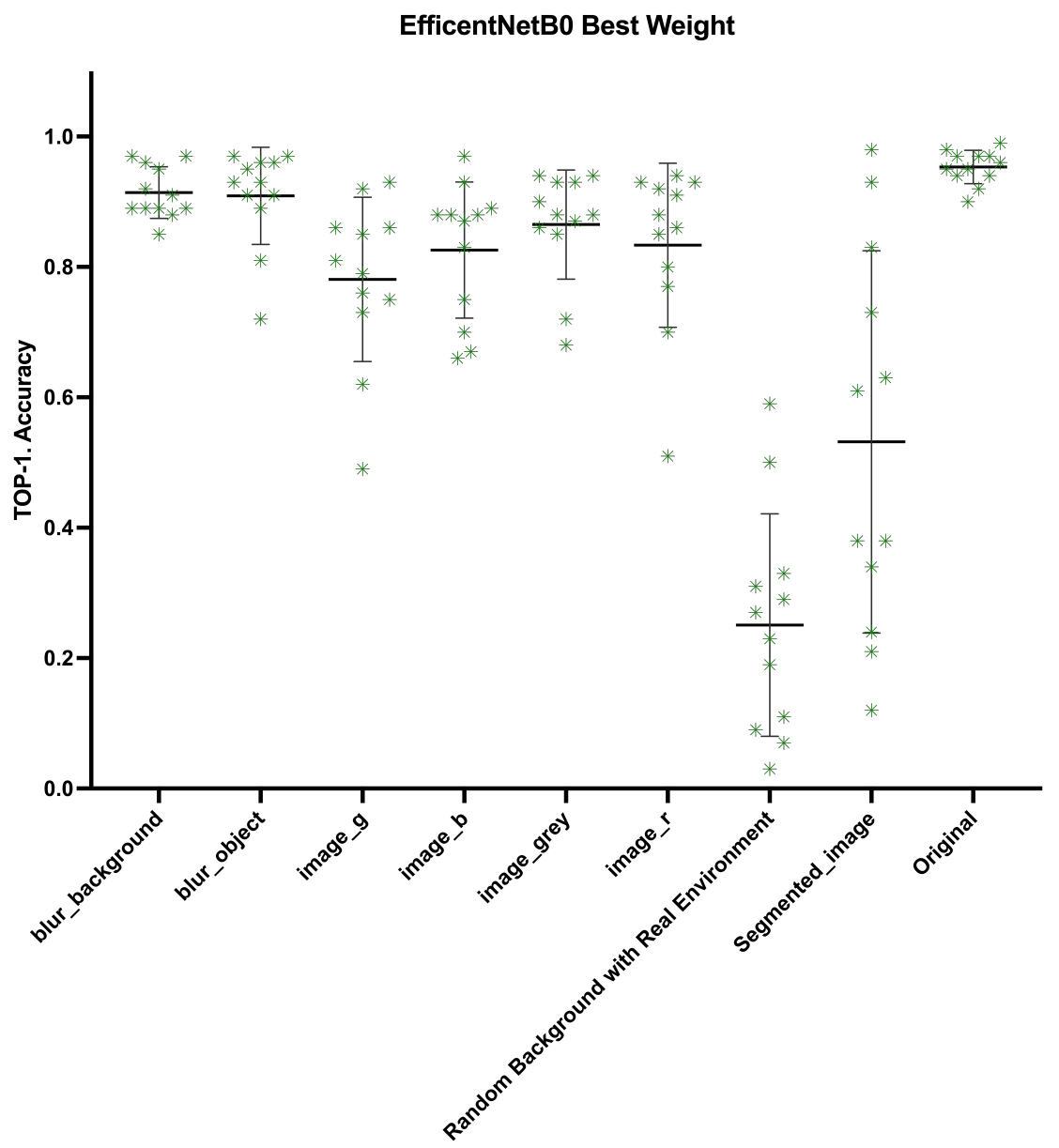}
        \\ Top-1 Accuracy on EfficientNetB0 \cite{tan2019efficientnet}
    \end{minipage}
    \hfill
    \begin{minipage}{0.32\linewidth}
        \centering
        \includegraphics[width=1.\linewidth]{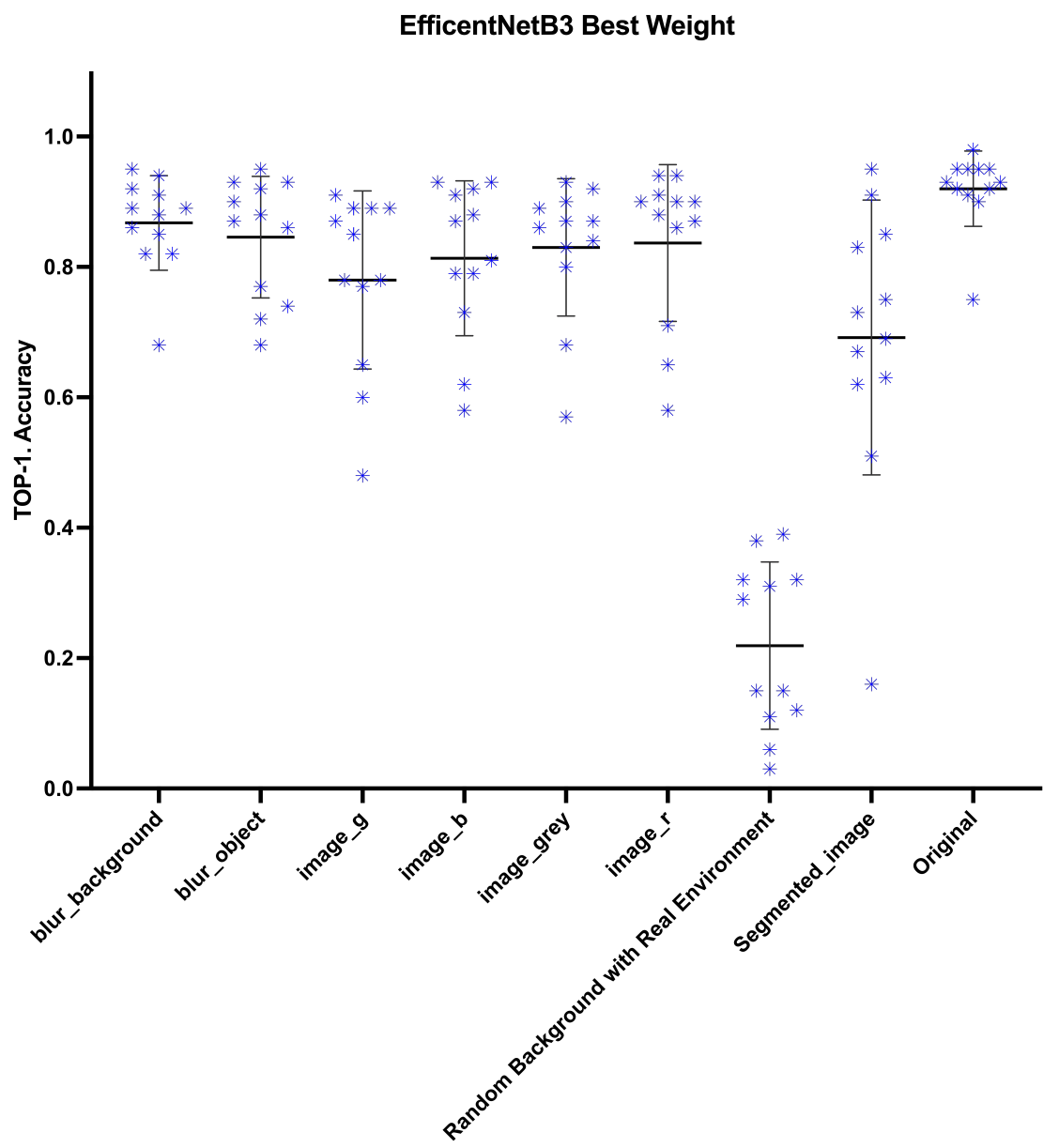}
        \\ Top-1 Accuracy on EfficientNetB3 \cite{tan2019efficientnet}
    \end{minipage}
    \hfill
    \begin{minipage}{0.32\linewidth}
        \centering
        \includegraphics[width=1\linewidth]{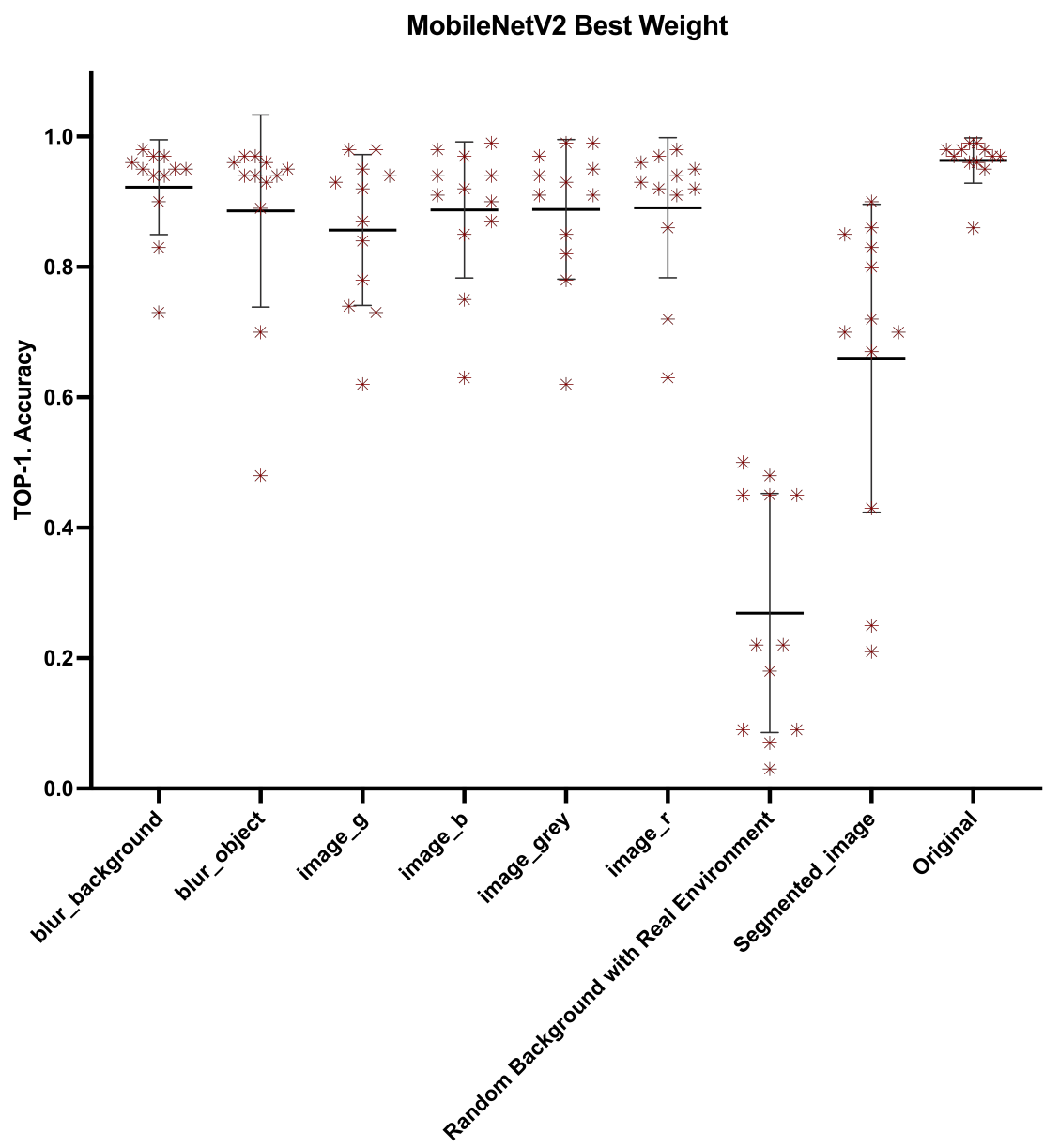}
        \\ Top-1 Accuracy on MobileNetV2 \cite{sandler2018mobilenetv2}
    \end{minipage}
    \hfill

    \vspace{0.5cm} 
    
    \begin{minipage}{0.32\linewidth}
        \centering
        \includegraphics[width=1\linewidth]{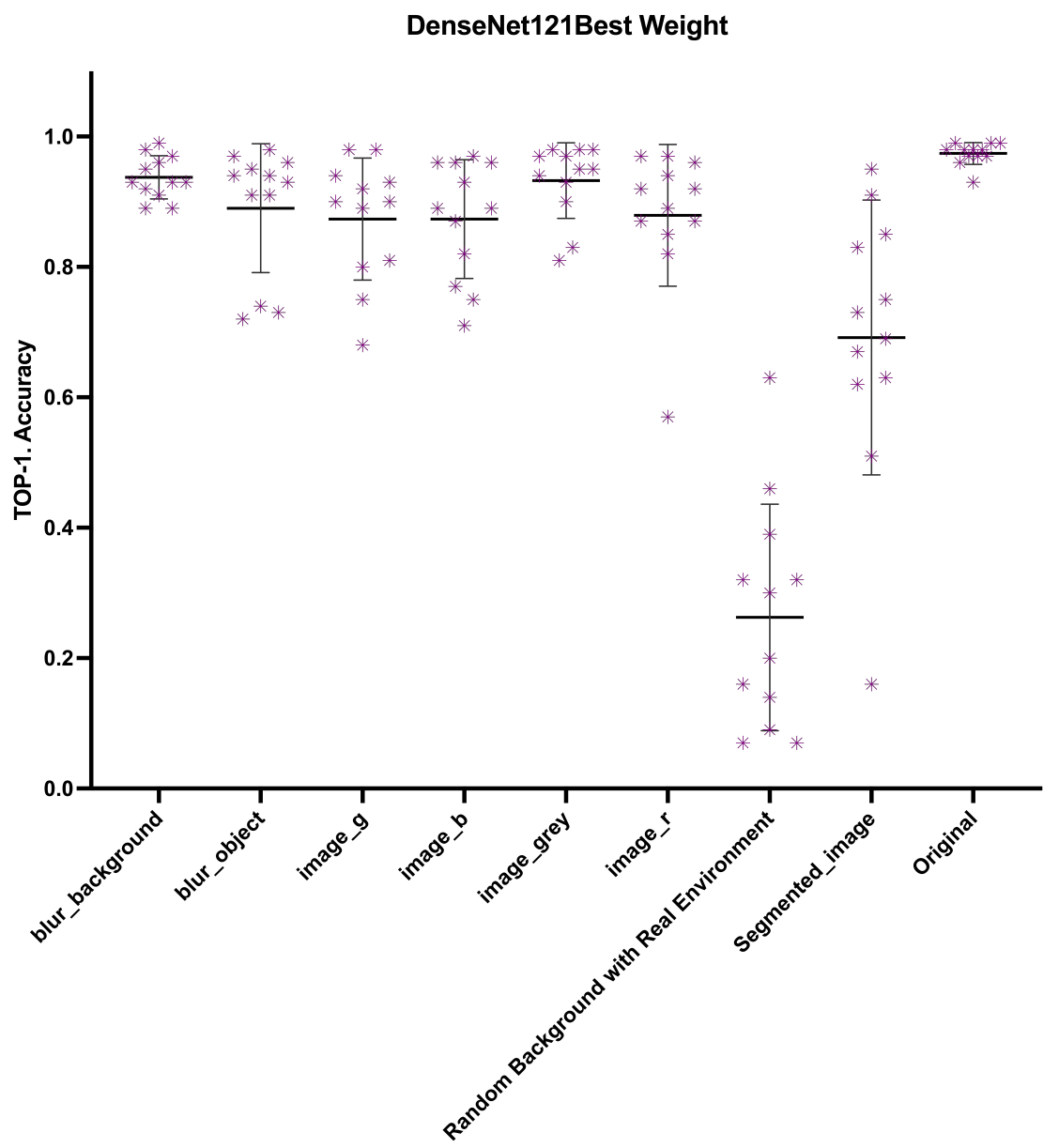}
        Top-1 Accuracy on DenseNet121 \cite{huang2017densely}
    \end{minipage}
    \hfill
    \begin{minipage}{0.32\linewidth}
        \centering
        \includegraphics[width=1\linewidth]{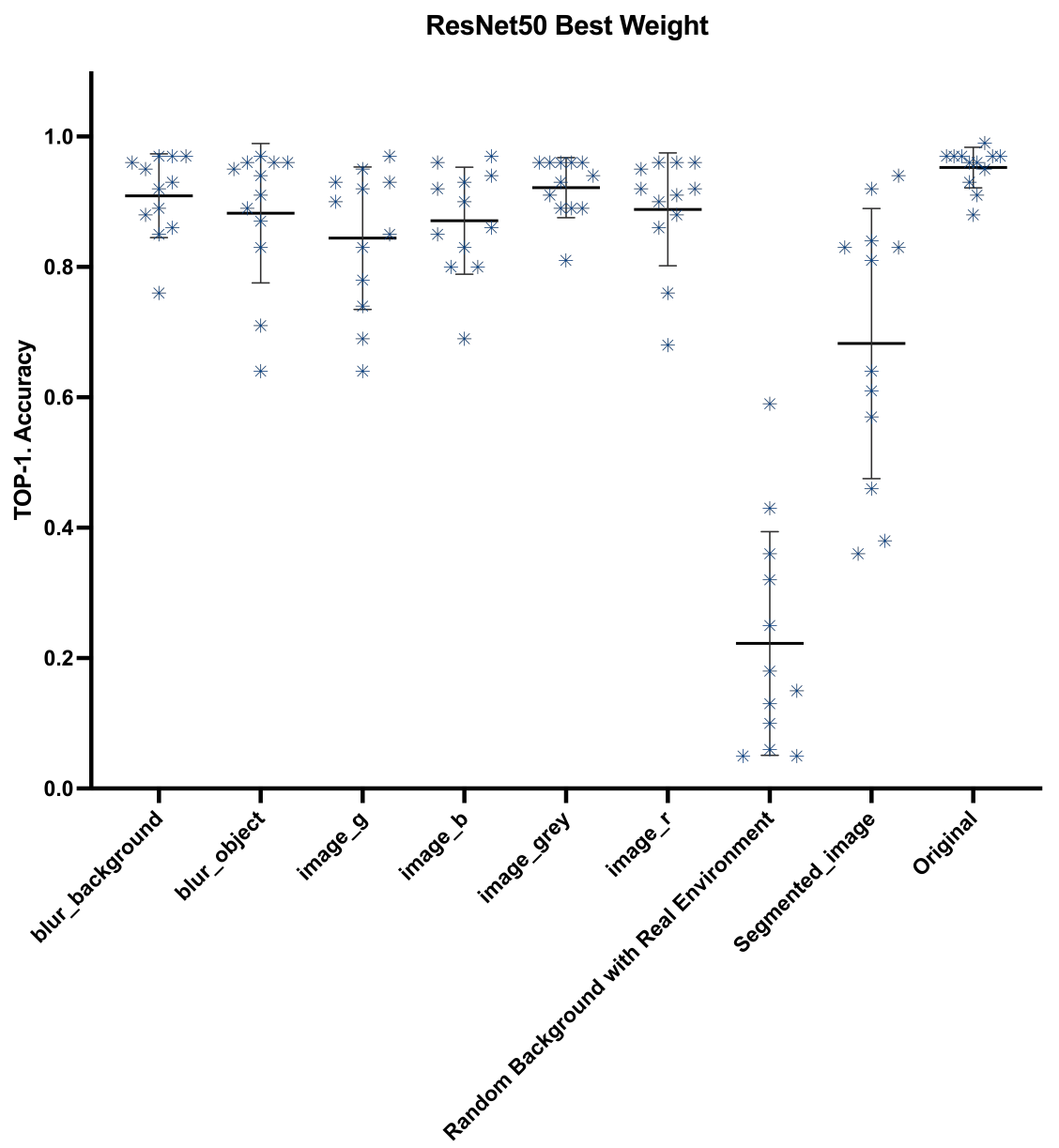}
        Top-1 Accuracy on ResNet50 \cite{he2016deep}
    \end{minipage}
    \hfill
    \begin{minipage}{0.32\linewidth}
        \centering
        \includegraphics[width=1\linewidth]{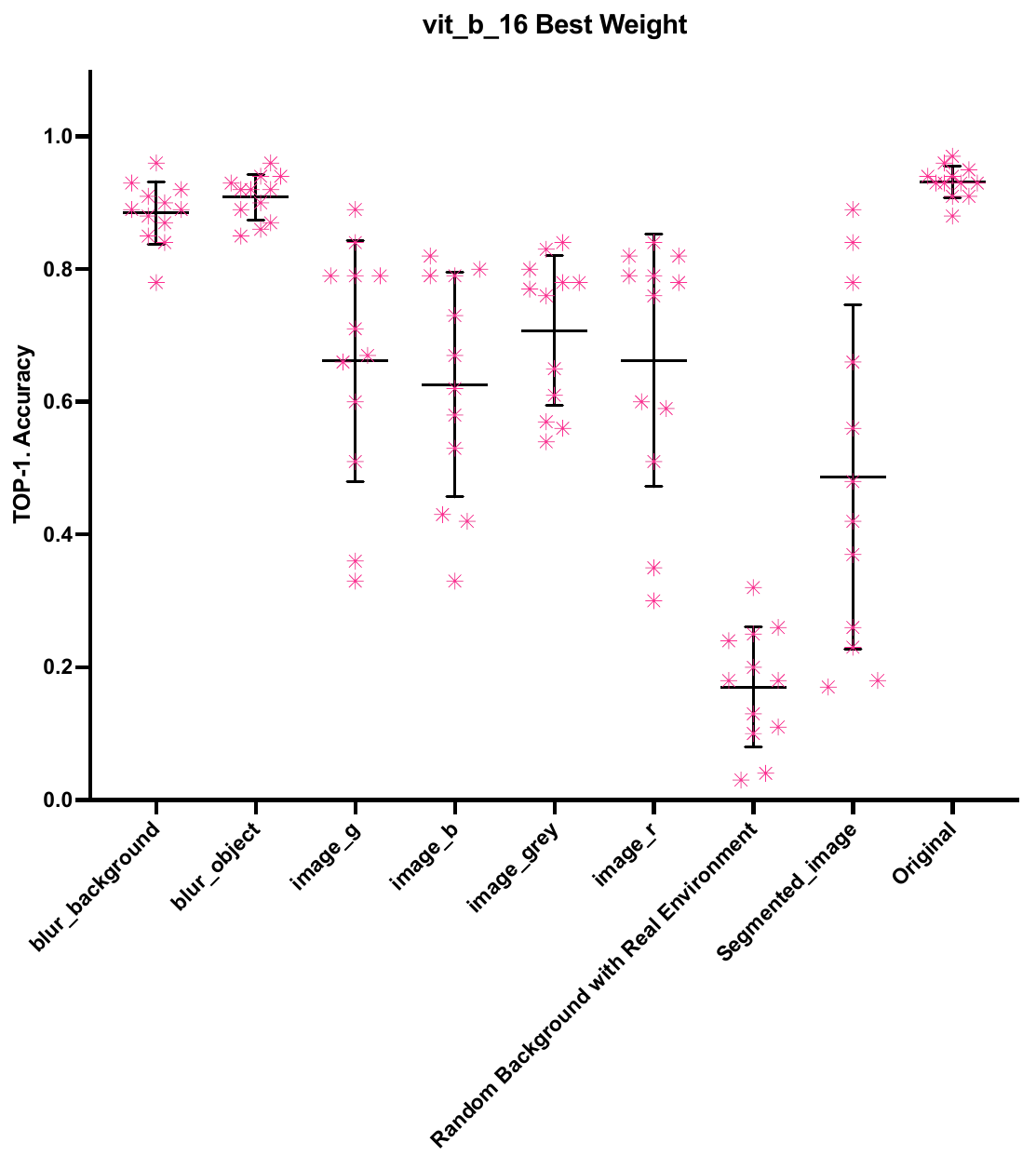}
        Top-1 Accuracy on ViT \cite{dosovitskiy2020image}
    \end{minipage}
    
    \caption{TOP-1 Accuracy for SOTA models pre-trained on IMAGENET original formal images and tested on XIMAGENET-12 different background scenarios. }
    \label{fig:accdrop}
\end{figure*}


\textbf{Performance Evaluation of Multiple Linear Regression.} Here, we further examine the changes in model performance across scenarios (EX1), using multiple linear regression analyses. In addition to addressing classification tasks, we extend our investigation to include segmentation tasks. 

As shown in Table \ref{tab:CombinedTable} of multiple linear regression analyses, we consider these three variables: Model Name, Scenarios, and Image Class. We use the P-value to indicate the confidence of our results and use the P-value summary as an auxiliary indicator of the P-value.

Specifically, the coefficient of Estimate for Segmentation Model Name[dpt-vitb16] is -0.1999, indicating that, compared to the reference segmentation Intercept model (deeplabv3plus-r50-d8~\cite{chen2018encoder}), model[dpt-vitb16] is associated with a decrease of 0.1999 in segmentation accuracy. With most of the SOTA Visual segmentation model Accuracy decrease from -0.05 to -0.22 and the base Intercept model only achieve 0.6672 Accuracy with (P $<$ 0.0001), it further verified that our benchmark could serve a valuable tool for measuring (SOTA) segmentation models performance in segmenting complex shapes or detecting detailed area in AI-generated background images (with Image Scenarios AI generated background leads to accuracy decrease by 0.14). 

Besides, we can see that models suffer a performance drop once the background changes as all Image Scenarios Accuracy dropped from -0.7078 (Random Background) to color change (image\_grey) -0.06517. 
Notably, the Classification(EX1) results in Table~\ref{tab:CombinedTable} also indicate \itshape \textbf{foreground class also play an important role for content reasoning}\upshape, since Image Class[1,4,5] will lead to Accuracy increase in replacement of baseline Intercept Image Class[0], while other leads to decrease. 

The results of the regression analyses are presented in Table \ref{tab:CombinedTable}, confirming our hypotheses. \itshape \textbf{Our benchmark should present also a challenging task for SOTA segmentation models}\upshape. It serves as an effective tool for assessing model performance in segmenting complex shapes and detecting detailed areas within AI-generated background images.



\textbf{Accuracy Drop of SOTA Models.}  We show the accuracy variance of SOTA visual models in Figure~\ref{fig:accdrop}. Notably, we observed that the presence of ``Random Background" had the most substantial adverse impact on accuracy, resulting in a significant decrease. Furthermore, the ``Segmented Image" scenario also exhibited a significant negative influence, leading to a decrease in accuracy.

\begin{figure}[t]
    \centering
    
    \begin{minipage}{0.43\textwidth}
        \centering
        \includegraphics[width=1.1\linewidth]{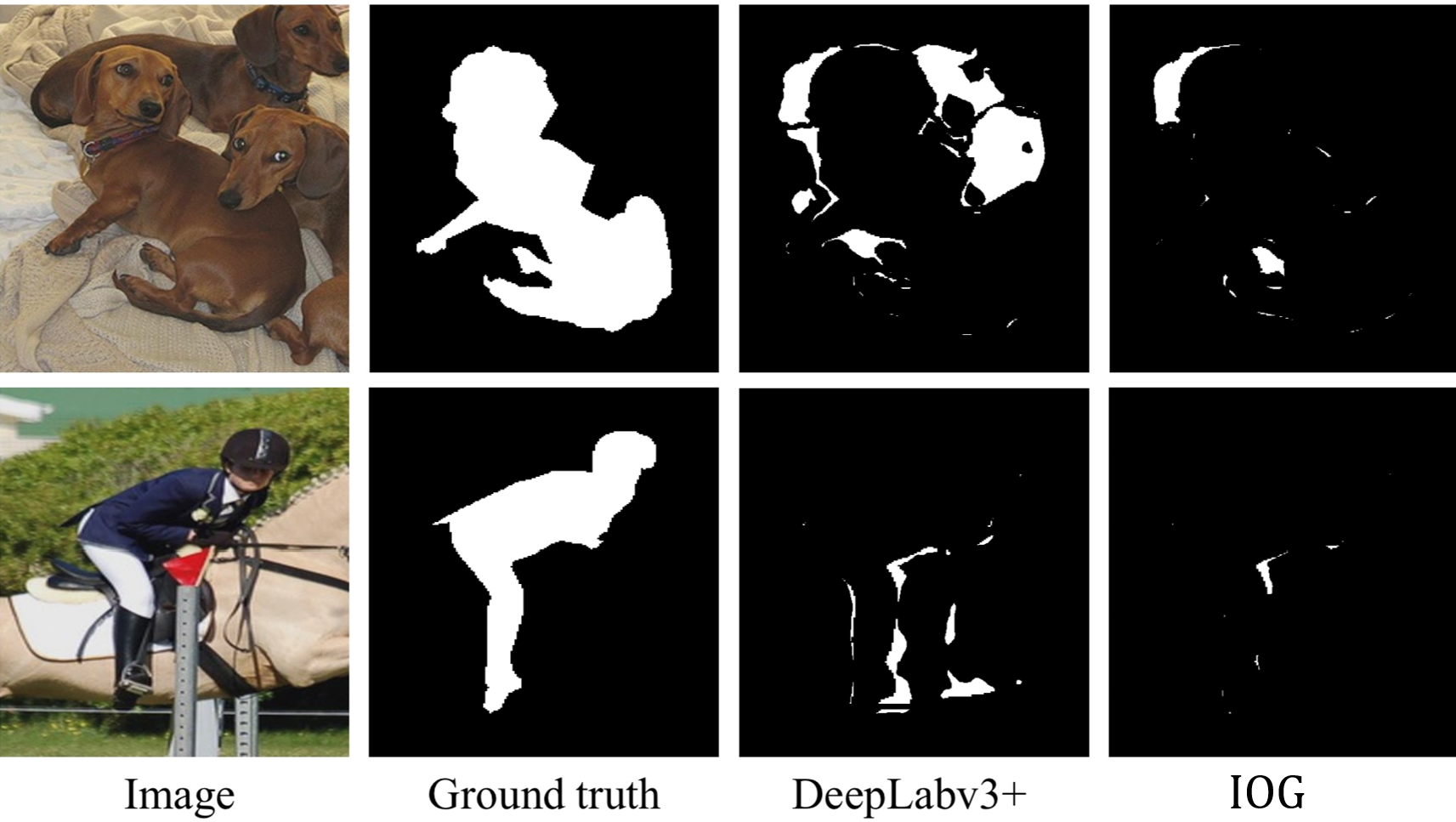}
        \caption*{IoG Benchmark Dataset}
    \end{minipage}%
    \hfill
    \hfill
    \begin{minipage}{0.44\textwidth}
        \centering
        \includegraphics[width=1.1\linewidth]{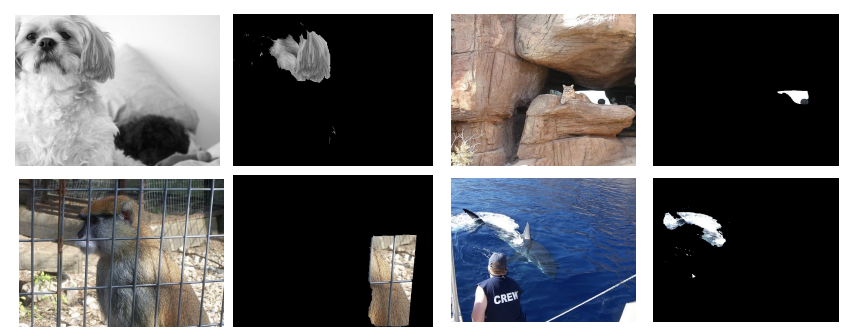}
        \caption*{ImageNet-9 Dataset}
    \end{minipage}%
    
    \hfill
    \hfill
    
    \begin{minipage}{0.45\textwidth}
        \centering
        \includegraphics[width=1.1\linewidth]{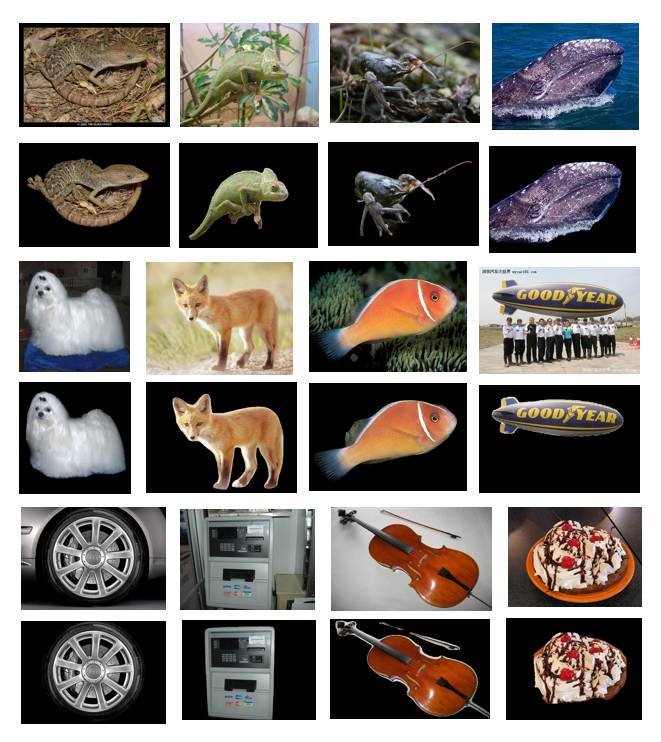}
        \caption*{XIMAGENET-12 Dataset}
    \end{minipage}%
    \hfill
    \caption{Additional Related Works and Explicit Comparisons. As can be seen, the semantic annotation of XIMAGENET-12 dataset is much more precise than the other datasets.}
    \label{fig:drawback1}
\end{figure}

 \subsection{Qualitative Results}
\label{gen_insti}
 We compare the semantic annotations of our XIMAGENET-12 with the IOG benchmark dataset~\cite{DBLP:conf/ijcai/ZhuXY17} and ImageNet-9 dataset~\cite{xiao2020noise} in Figure~\ref{fig:drawback1}. As can be seen, the semantic labels of XIMAGENET-12 are much more precise than the others. We consider that due to the more precise separation of foreground and background, we can conduct a more reliable evaluation and analysis of the model robustness. For example, the sup-optimal annotation of ImageNet-9~\cite{xiao2020noise} leads to a misleading claim that removing the background negatively impacts test accuracy. In contrast, we argue that poor segmentation quality, particularly with minimal foreground remaining, hampers the performance of recognition. We believe that our dataset can perform as a high-quality dataset for analysis of domain adaptation/generation. 
 
\begin{figure}[t]
    \centering
    
    \begin{minipage}{0.45\textwidth}
        \centering
        \includegraphics[width=1.0\linewidth]{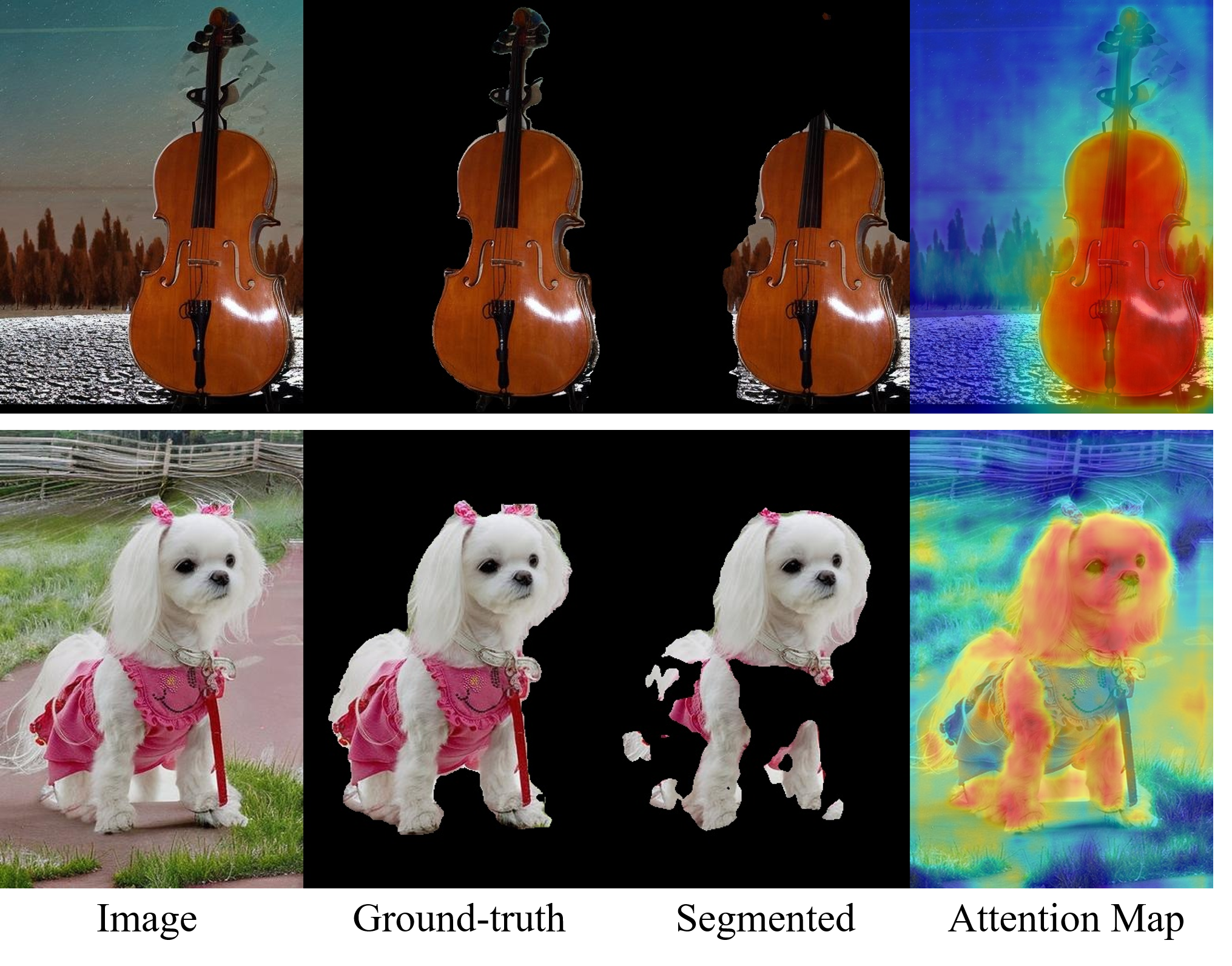}
    \end{minipage}%
    \hfill
    \caption{Attention Map of SOTA segmentation models. ``pspnet \cite{zhao2017pyramid} r50-d8" achieves a MIoU of 0.562 in our dataset. Our dataset is also enabled to tackle the intricacies of segmenting objects within AI-generated backgrounds, offering a substantial improvement in human-labeled ground truth quality compared to the original ImageNet \cite{deng2009imagenet}, where only image labels are included. Moreover, it proves to be a valuable resource for identifying AI-generated images on the internet, showcasing its versatility and significance in contemporary computer vision research.}
    \label{fig:type3}
\end{figure}

 We show the segmentation and attention map of SOTA segmentation models on XIMAGNET-12 in Figure~\ref{fig:type3}. As can be seen, those segmentation models do not show satisfying performance. For example, the clothes of the dog has not been recognized. This indicates that our XIMAGNET-12 is a challenging dataset for the segmentation task. 
 

\section{DISCUSSION AND CONCLUSION}
\label{gen_inst}

In this work, we introduce an explainable visual benchmark dataset, XIMAGENET-12, to evaluate the robustness of visual models. XIMAGENET-12 consists of six diverse scenarios, such as overexposure, blurring, color changes, etc., to simulate real-world situations. We further develop a robustness score to investigate the model performance under various conditions. From the experiments, we conclude the following interesting findings: 

1) Different scenarios influence visual models in different degrees, and randomly substituting the background leads to the most severe performance drops.

2) Models trained and tested with well-segmented foregrounds tend
to perform well even if the backgrounds are missing.

3) A model with higher accuracy is not necessarily more stable.

We expect the XIMAGENET-12 dataset will empower researchers to thoroughly evaluate the robustness of their visual models under challenging conditions. In future work, we will show how XIMAGENET-12 can serve as a high-quality dataset for more visual applications such as semantic segmentation tasks and domain adaptation/generation tasks.

{
    \small
    \bibliographystyle{ieeenat_fullname}
    \bibliography{main}
}
\clearpage
\maketitlesupplementary

\section{Appendix}
\label{sec:rationale}

In this section we provide the supplementary compiled together with the main paper includes:
\begin{itemize}
\item The illustration of how we use Multiple Linear Regression to verify our hypothesis: from raw data input, for example, in GraphPad Prism, to interpreting examples and residual plots, etc;
\item The training details (Accuracy and Loss Plots) and hyperparameters within scenarios and cross-scenario experiments, including diffusion metrics for evaluations, density maps of State-of-the-Art accuracy drop (e.g., referring to our particular experiment, EX1, EX2);
\item The ablation study addresses the industry pain points, illustrating robust model selection for challenging scenarios, particularly due to factors such as background variations, camera shifts, color changes, and lighting conditions;
\item Sample image of our XIMAGENEt-12 AI-generated image, comprising 12,248 images for AI-generated scenarios using the latest Stable Diffusion XL model and involving uniform promotion and manual selection and filtering.

\end{itemize}


\begin{figure*}[h]
    \centering
    \begin{minipage}{0.45\textwidth}
        \centering
        \includegraphics[width=1\linewidth]{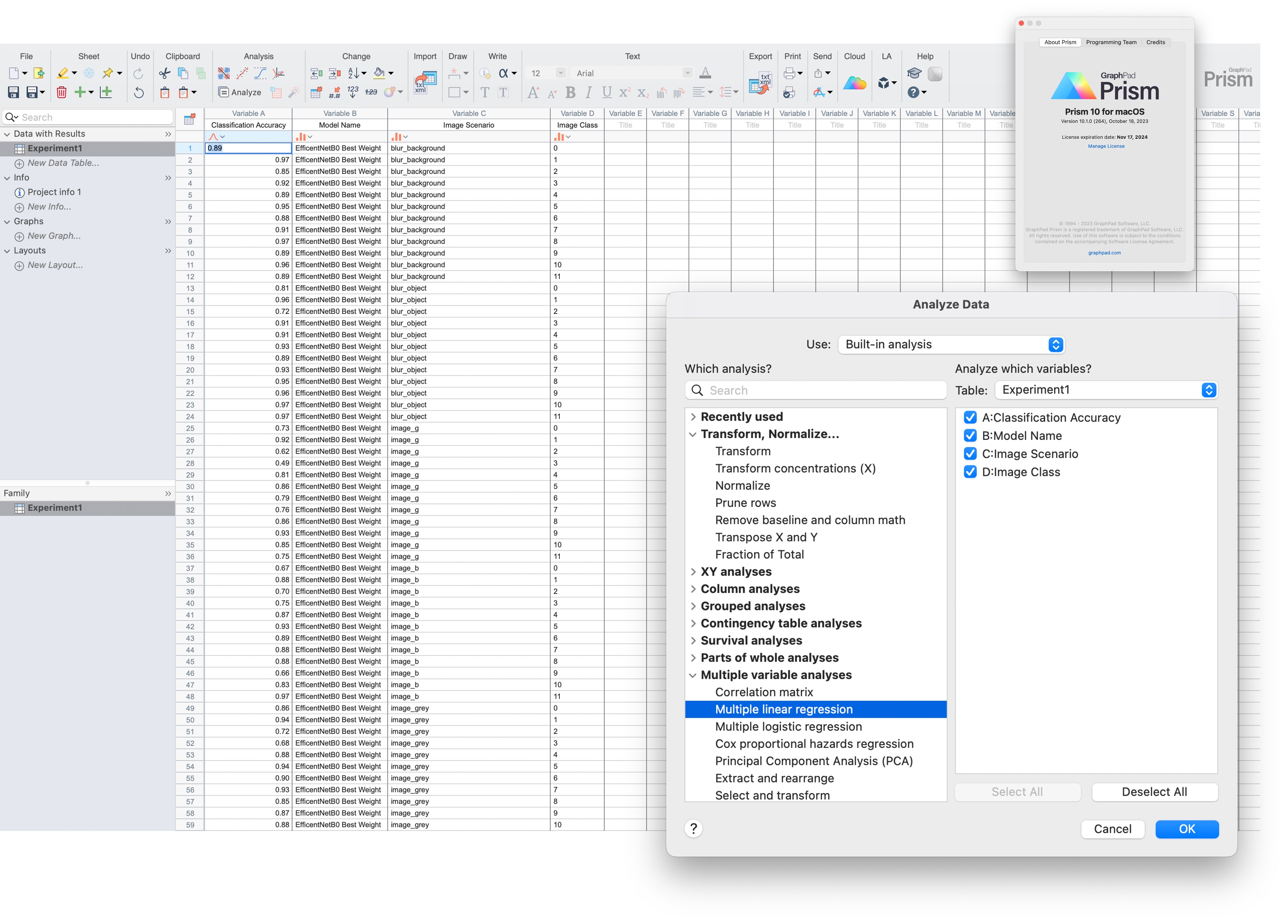}
        Define the raw data type and variable into statistic software (GraphPad Prism)
    \end{minipage}%
    \hfill
    \begin{minipage}{0.45\textwidth}
        \centering
        \includegraphics[width=1\linewidth]{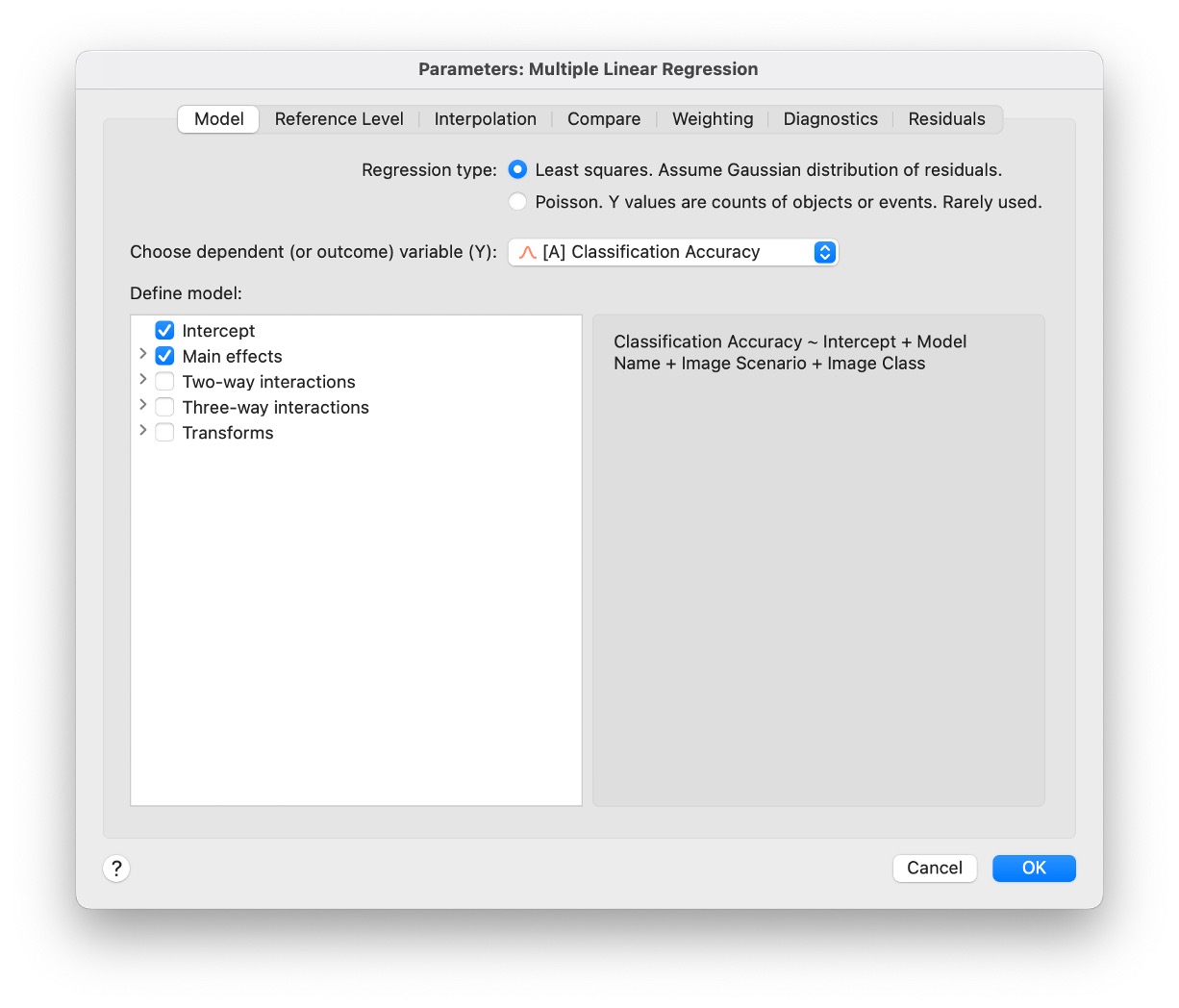}
        Choose Regression type and define the base independent variables (Model, Image Scenario, Image Class) 
    \end{minipage}%
    \hfill
    \begin{minipage}{0.45\textwidth}
        \centering
        \includegraphics[width=1\linewidth]{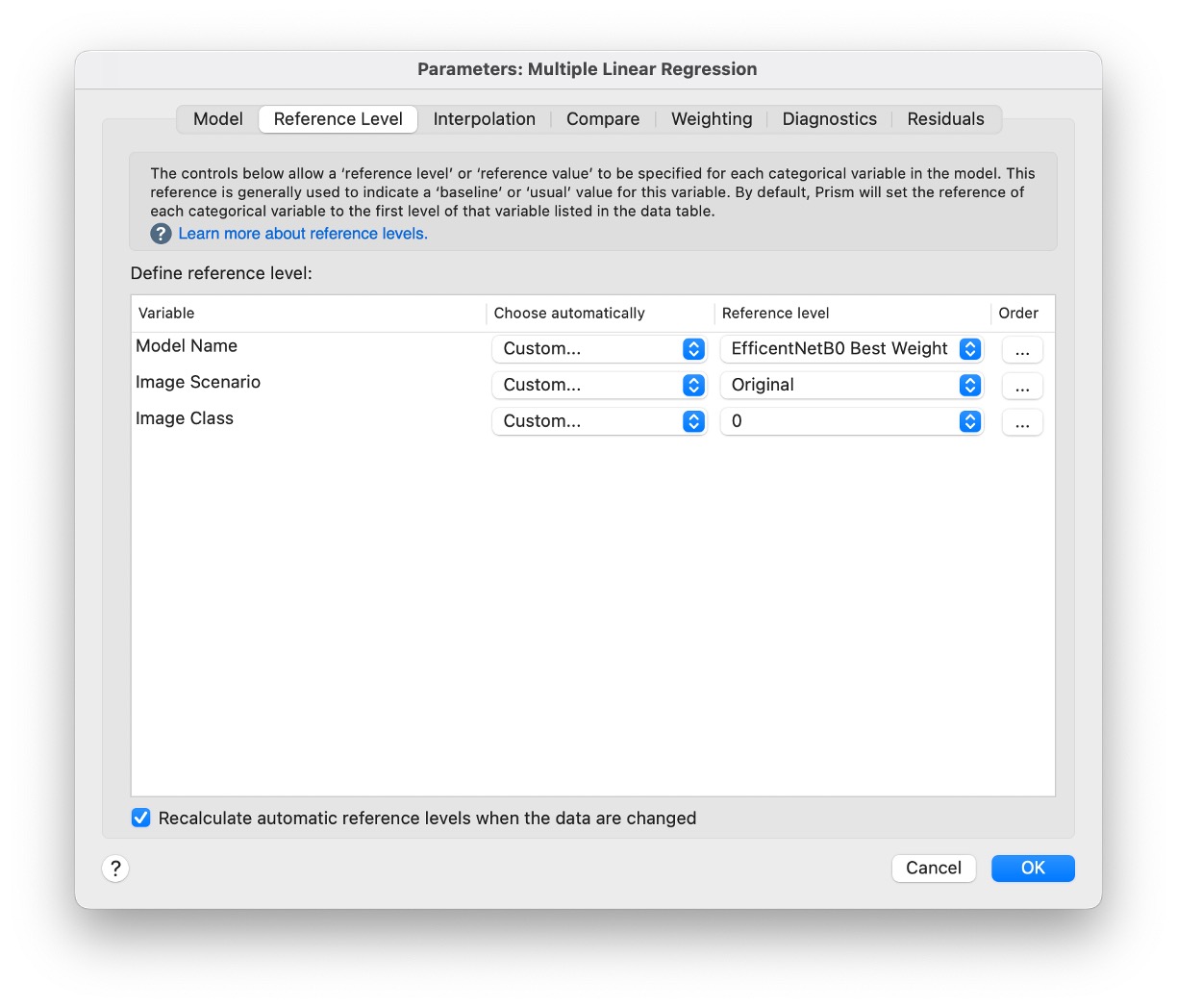}
        Select reference level for each independent variables (Model, Image Scenario, Image Class)
    \end{minipage}%
    \hfill
    \begin{minipage}{0.45\textwidth}
        \centering
        \includegraphics[width=1\linewidth]{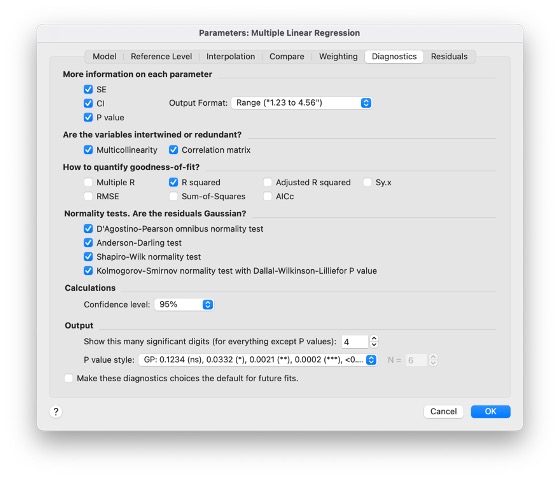}
        Set Parameters for Multiple Liner Regression, such as Confidence Level etc
    \end{minipage}%
    \hfill
    \begin{minipage}{0.45\textwidth}
        \centering
        \includegraphics[width=1\linewidth]{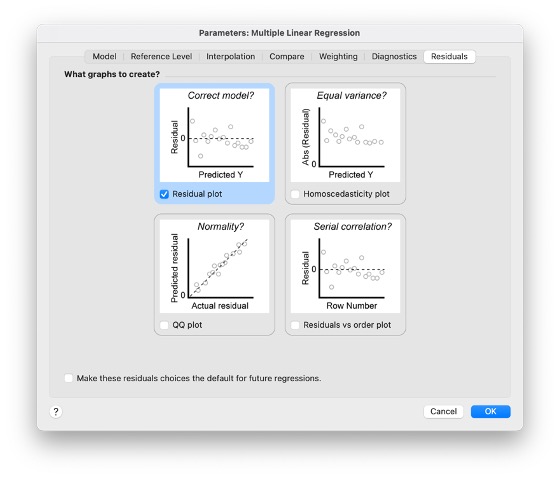}
        Create target residual plot graph for simulating the regression results
    \end{minipage}%
    \hfill
    \begin{minipage}{0.45\textwidth}
        \centering
        \includegraphics[width=1\linewidth]{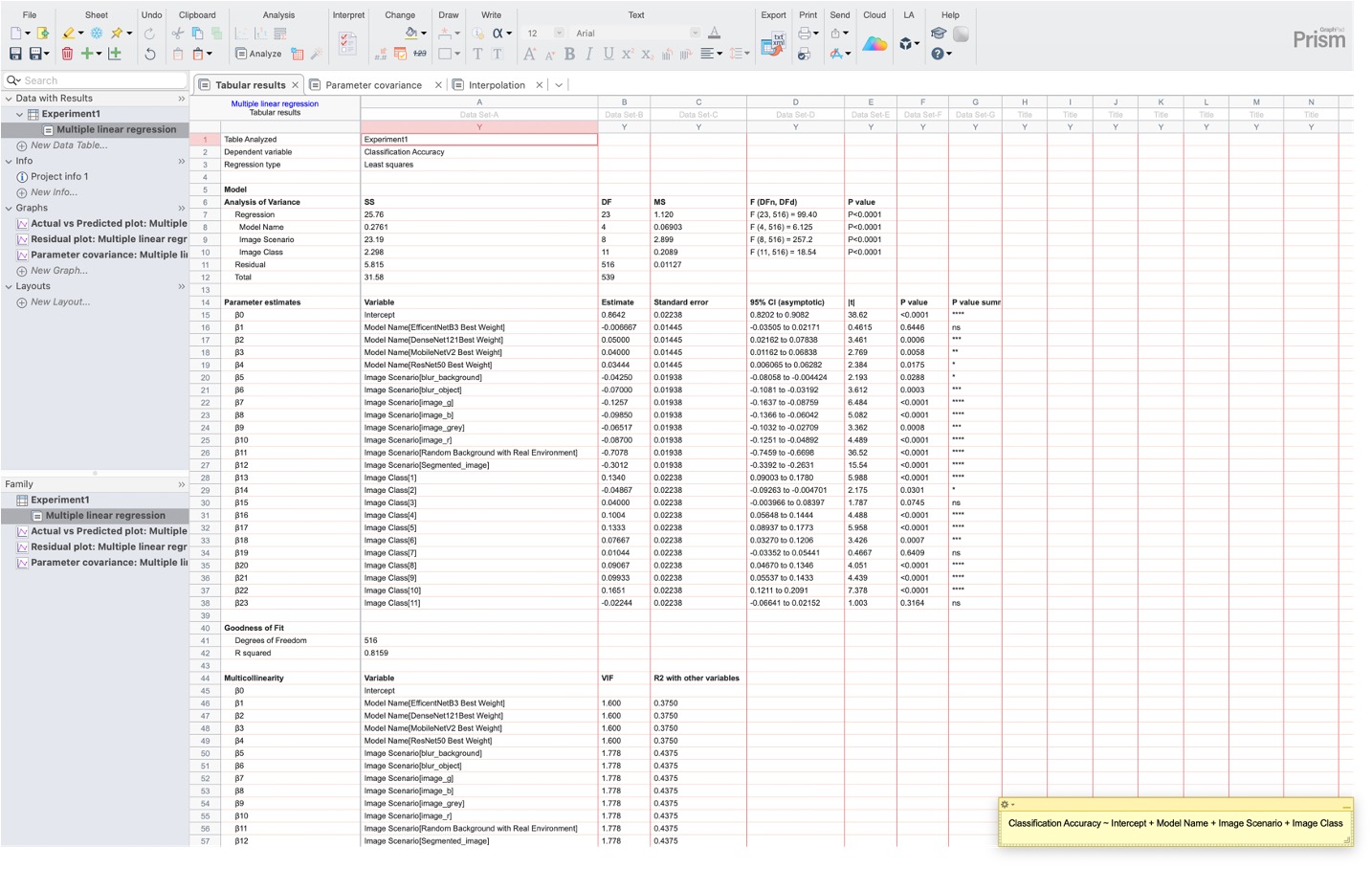}
         Generate the analyse and interpretation report includes Estimates and P Value for each variables
    \end{minipage}%
    \hfill
    \caption{Multiple Linear Regression Workflow and Example of Interpretations.}
    \label{fig: EX2}
\end{figure*}

\begin{figure*}[t]
    \centering
    \begin{minipage}{1\textwidth}
        \centering
            \includegraphics[width=1\linewidth]{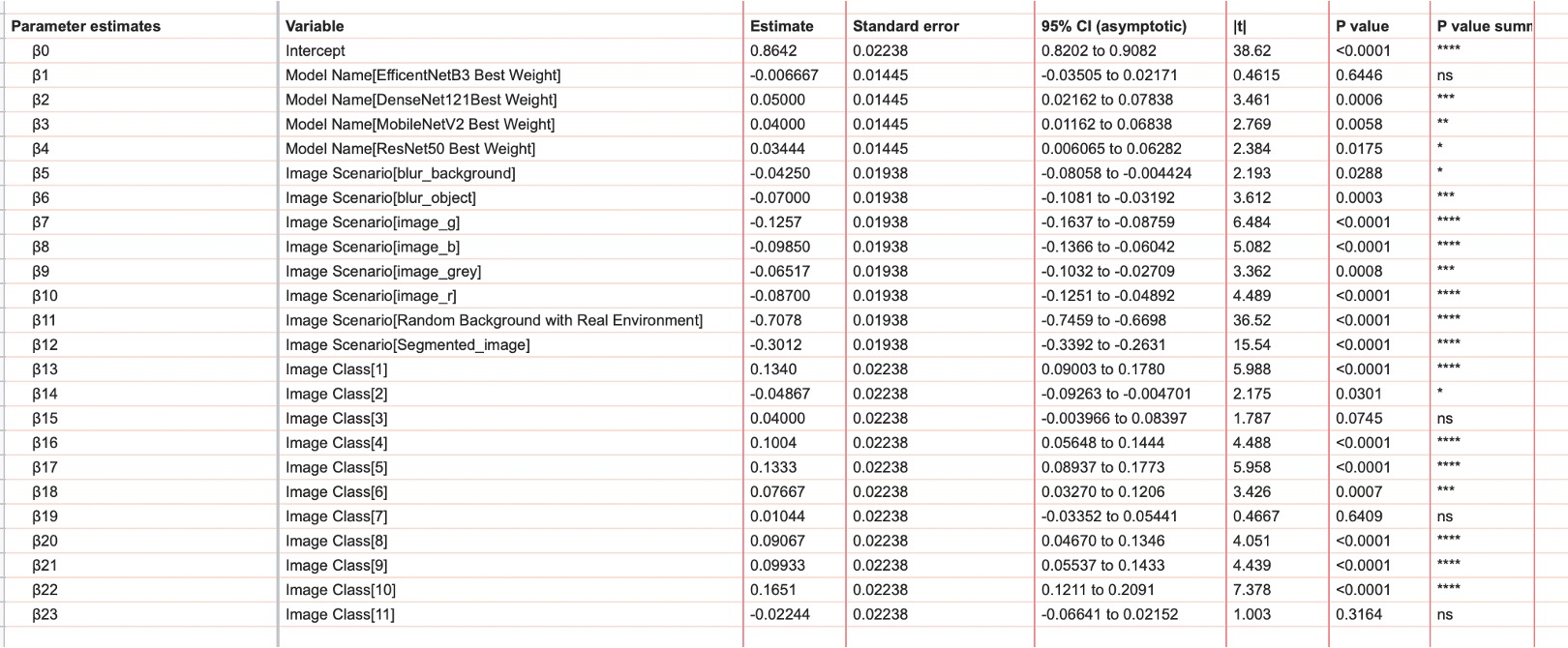} 
    \end{minipage}%
    \hfill
    \hfill
    \caption{Examples of Multiple Linear Regression Interpretations: (1) $\beta0$ (Intercept) estimate equal to 0.8642 means that the base classification accuracy when all predictors are at their reference levels is 86.42\%.  (2) $\beta2$ (Model Name [DenseNet121 Best Weight]) estimate equal to 0.05000, $P$ value 0.0006 means that the model [DenseNet121 Best Weight] increases the classification accuracy by 5.000\% when compared to the reference level the model [EfficentNetB0 Best Weight]. This effect is also statically significant ($P$ value \textless 0.05). (3) $\beta11$ (Image Scenario [Random Background with Real Environment]) estimate equal to -0.7078; $P$ value \textless 0.0001 means that this image scenario decreases the classification accuracy by 70.78\% when compared to the reference level the Image Scenario [Original] with a significant confidence. 
}
    \label{fig: bmw}
\end{figure*}

\begin{figure*}[t]
    \centering
    \begin{minipage}{0.24\textwidth}
        \centering
        \includegraphics[width=\linewidth]{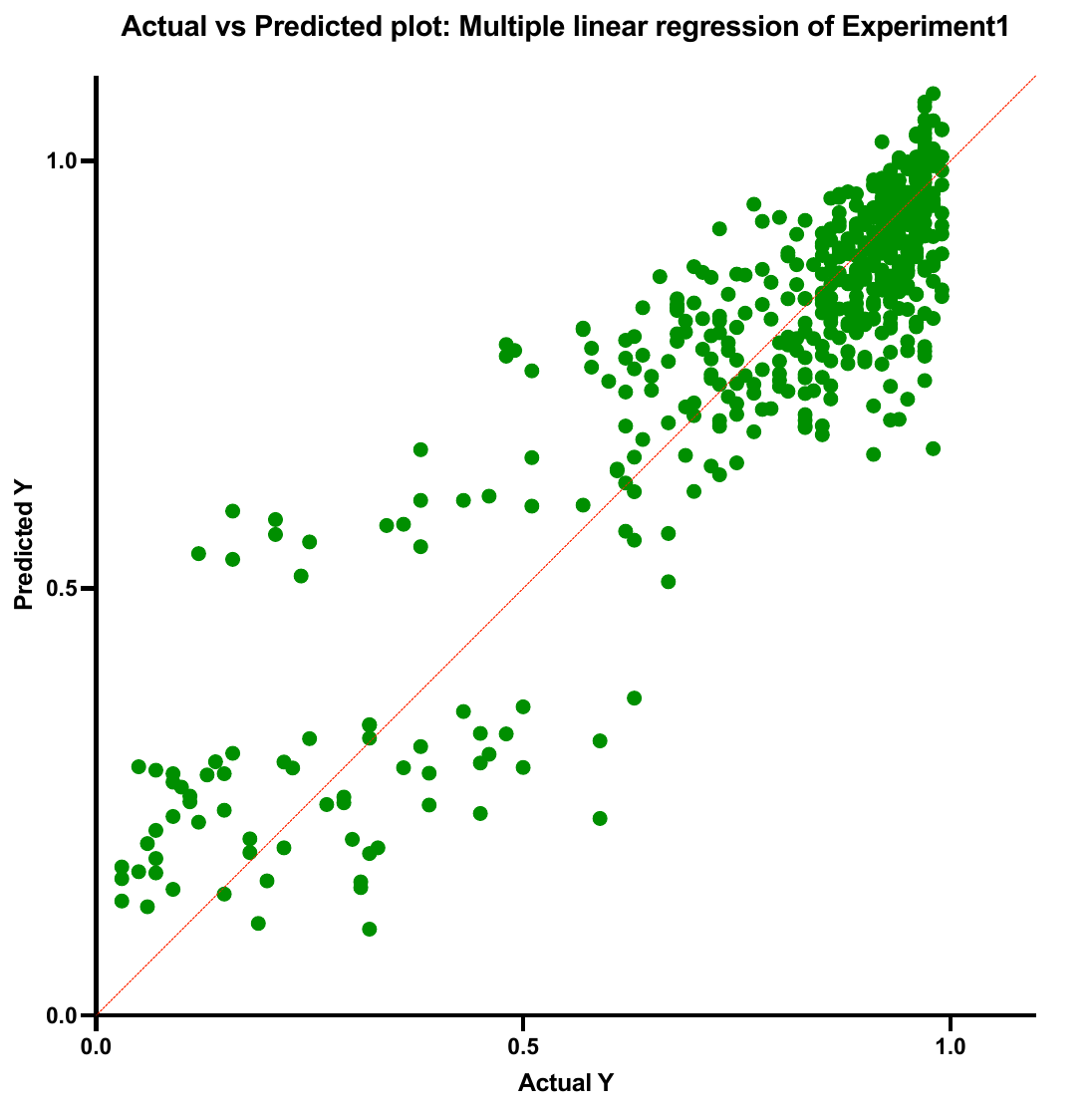}  
    \end{minipage}%
    \hfill
    \begin{minipage}{0.24\textwidth}
        \centering
        \includegraphics[width=\linewidth]{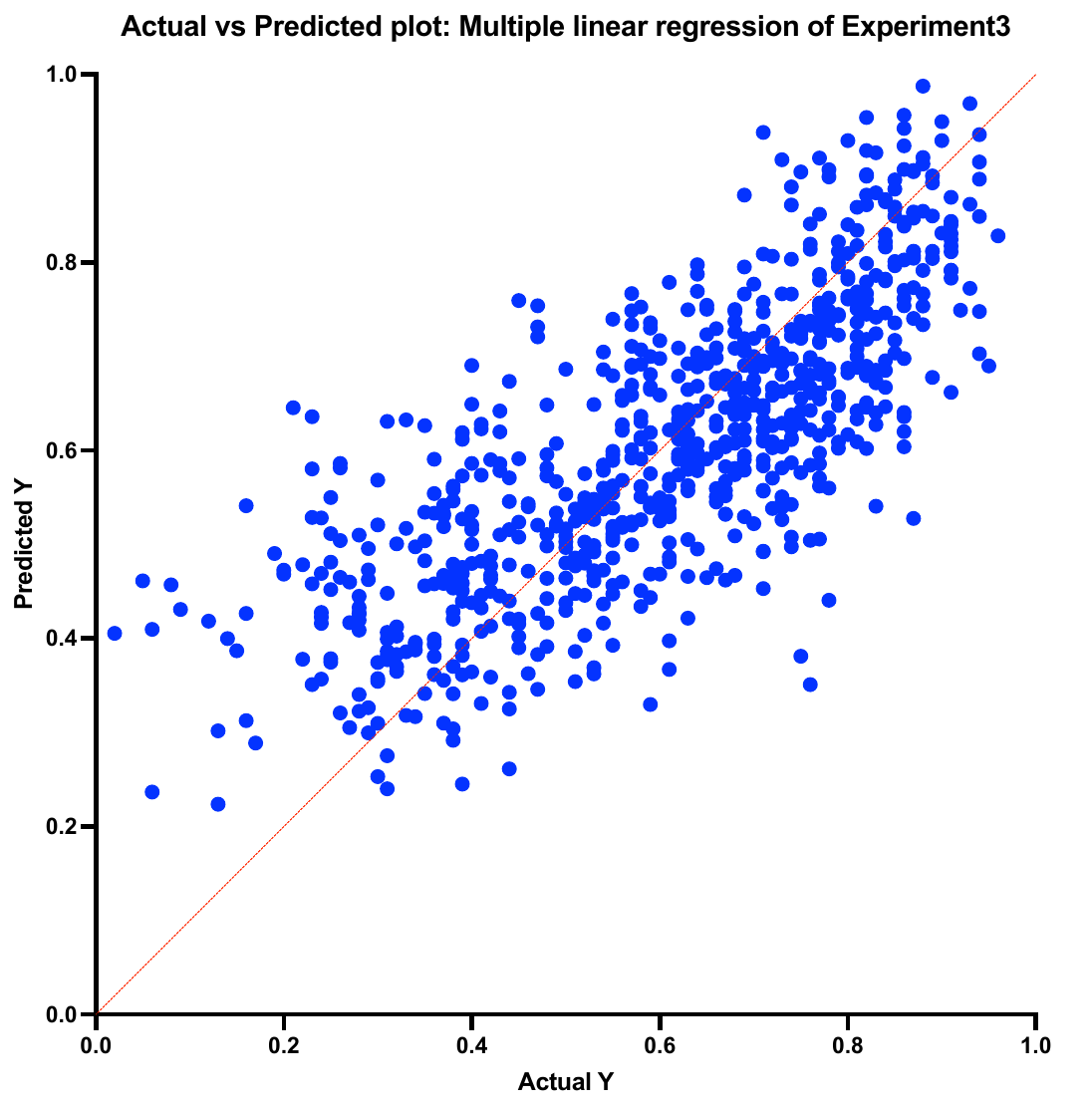}
    \end{minipage}%
    \hfill
    \begin{minipage}{0.24\textwidth}
        \centering
        \includegraphics[width=\linewidth]{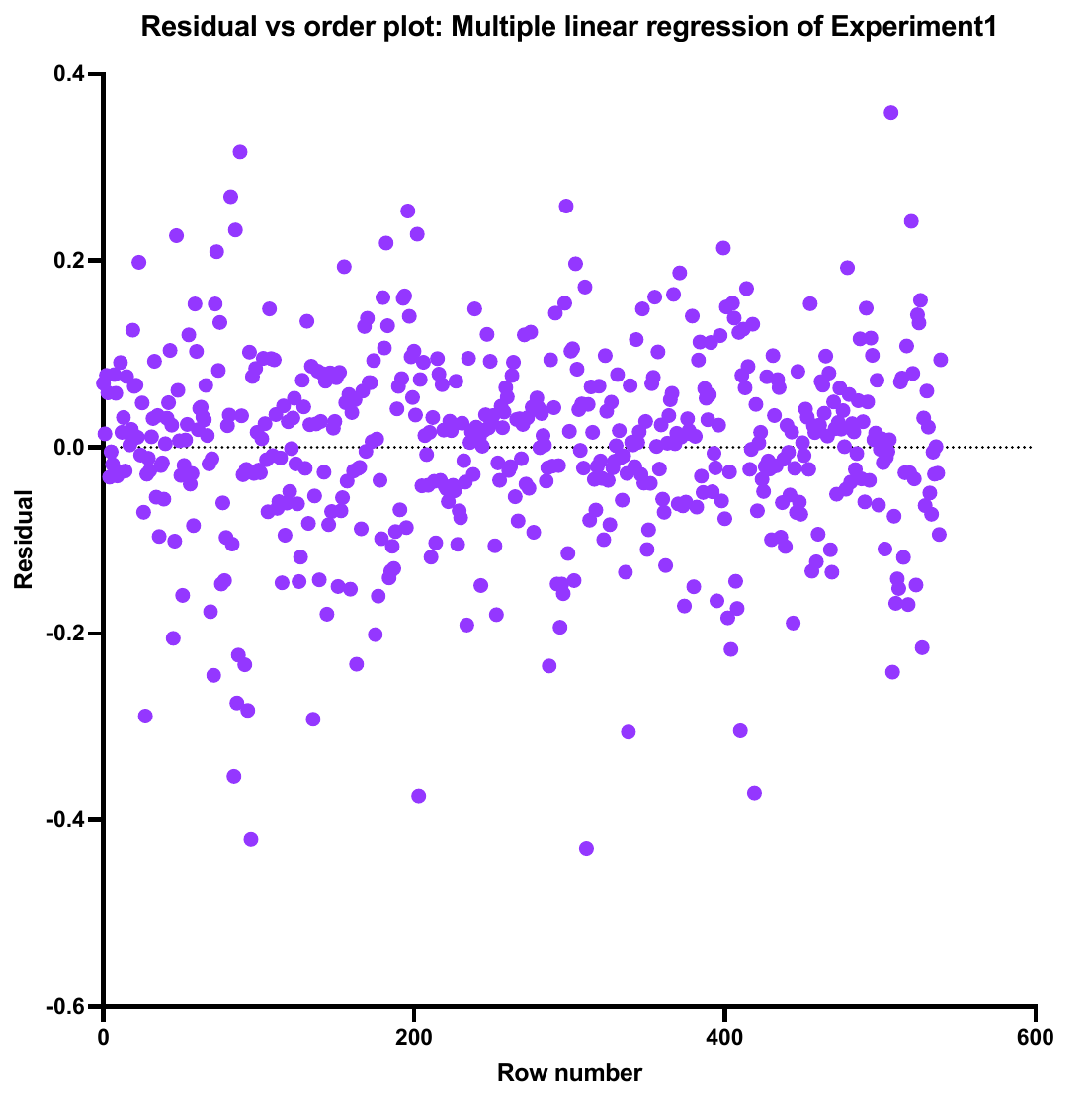}
    \end{minipage}%
    \hfill
    \begin{minipage}{0.24\textwidth}
        \centering
        \includegraphics[width=\linewidth]{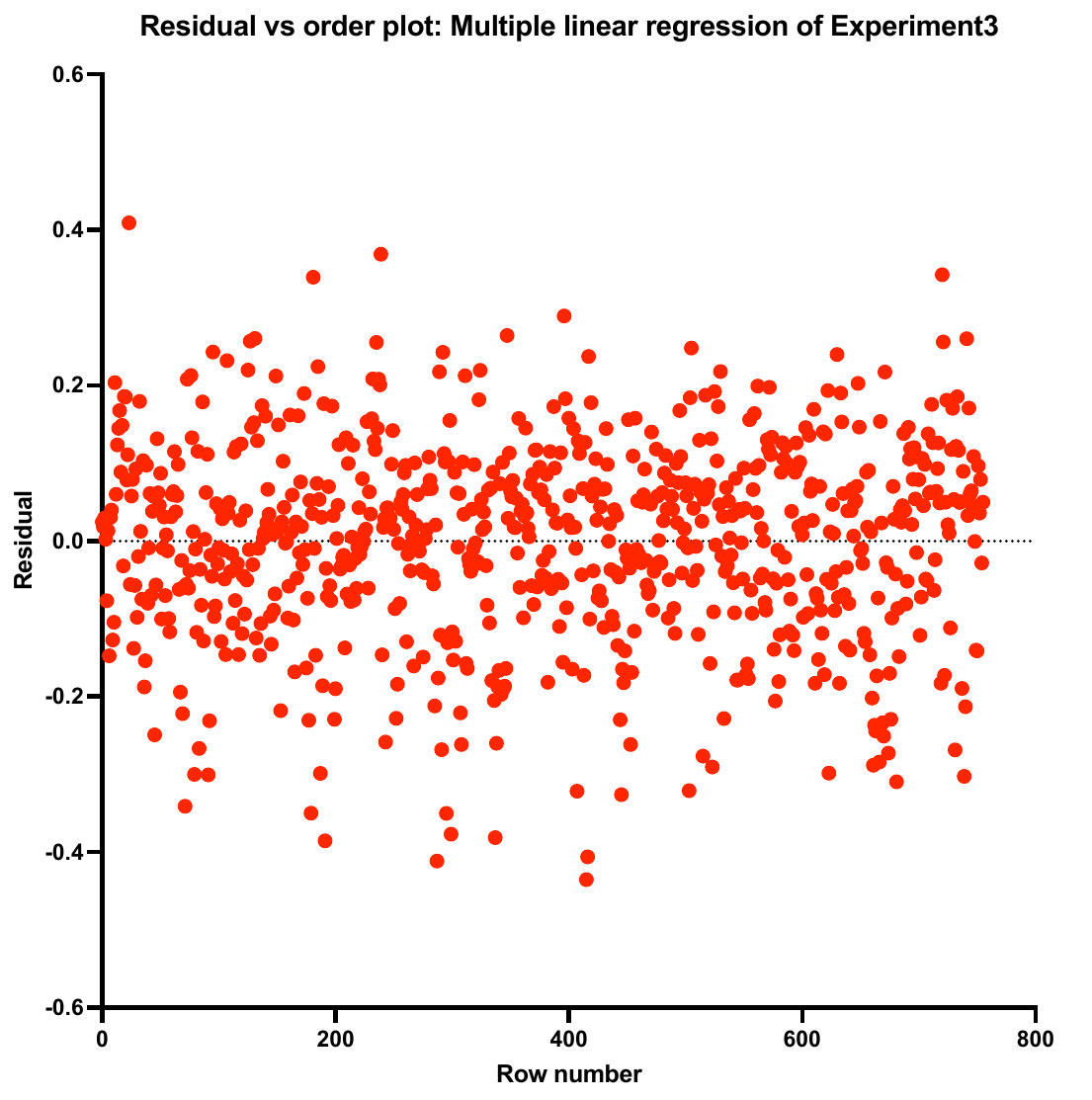}
    \end{minipage}%
    \hfill
    \caption{Multiple linear Regression  Residual Plot and QQ Plot of Ex1 (SOTA Model Classfication) and Ex3 (SOTA Segmentation) on Hypothesis Verification Process. These shows how two distributions (accuray point)' quantiles line up, with our theoretical distribution (e.g., the normal distribution) as the x variable (Scenarios, Image Object, Model Name) and regression model residuals as the y variable. If the points lie on or close to a 45-degree line, it means that the data follow the reference distribution closely, boosting confidence in the regression results.}
    \label{fig: Multi-linear}
\end{figure*}

\begin{figure*}[t]
    \centering
    \begin{minipage}{0.3\textwidth}
        \centering
        \includegraphics[width=\linewidth]{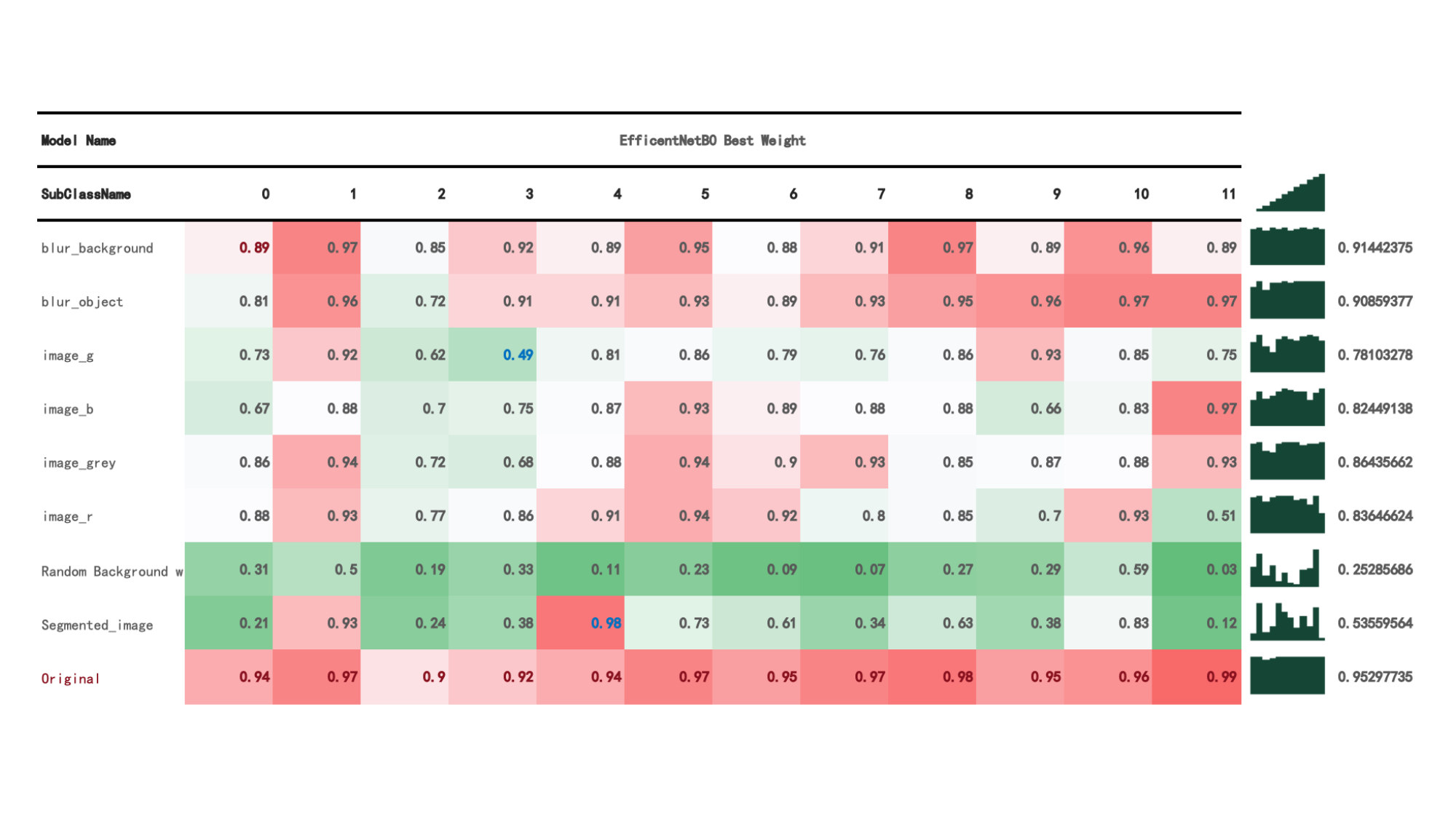}
        The EfficientNetB0 \cite{tan2019efficientnet} accuracy for each class
    \end{minipage}
    \hfill
    \begin{minipage}{0.3\textwidth}
        \centering
        \includegraphics[width=\linewidth]{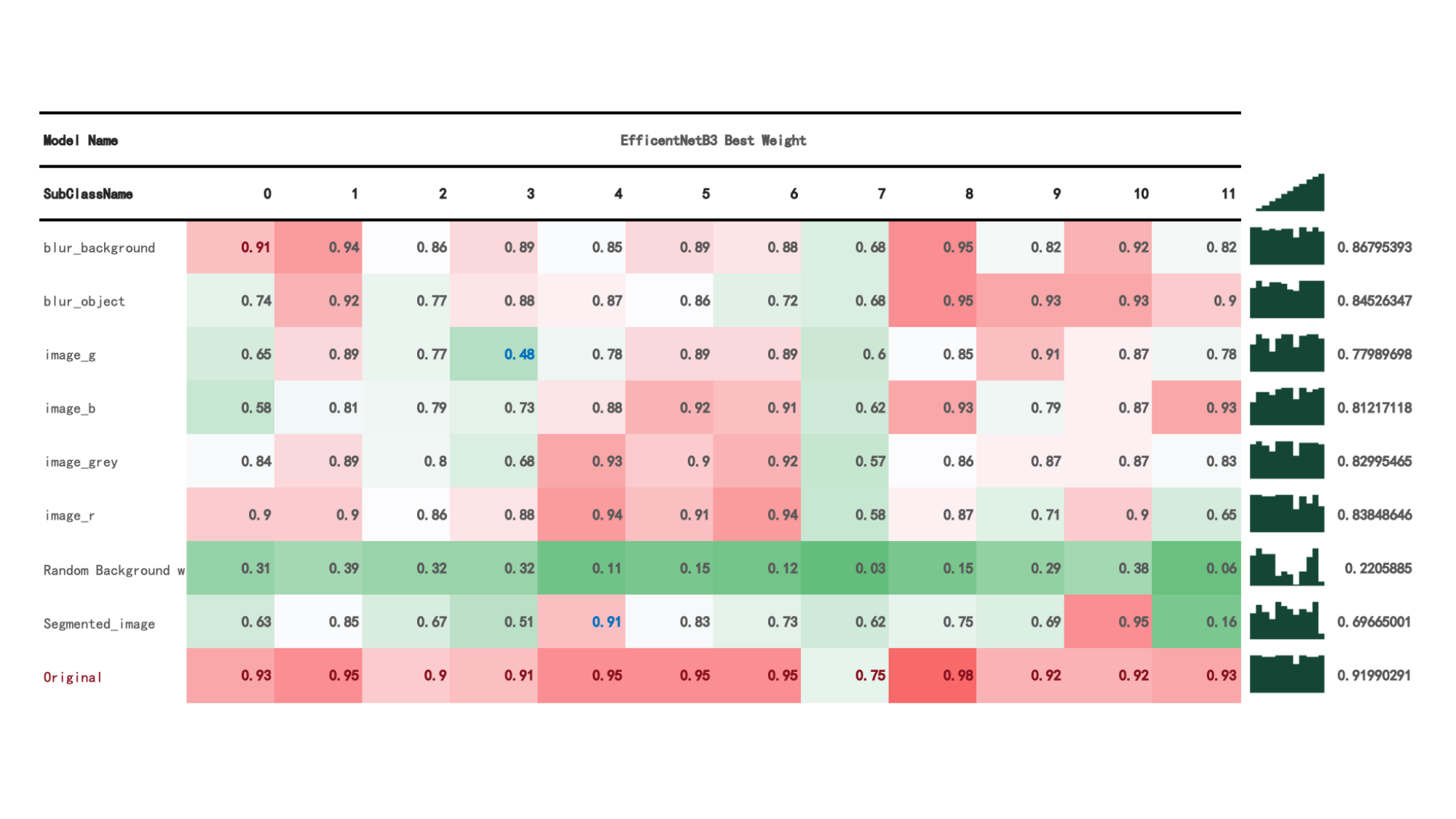}
        The EfficientNetB3 \cite{tan2019efficientnet} accuracy for each class
    \end{minipage}
    \hfill
    \begin{minipage}{0.3\textwidth}
        \centering
        \includegraphics[width=\linewidth]{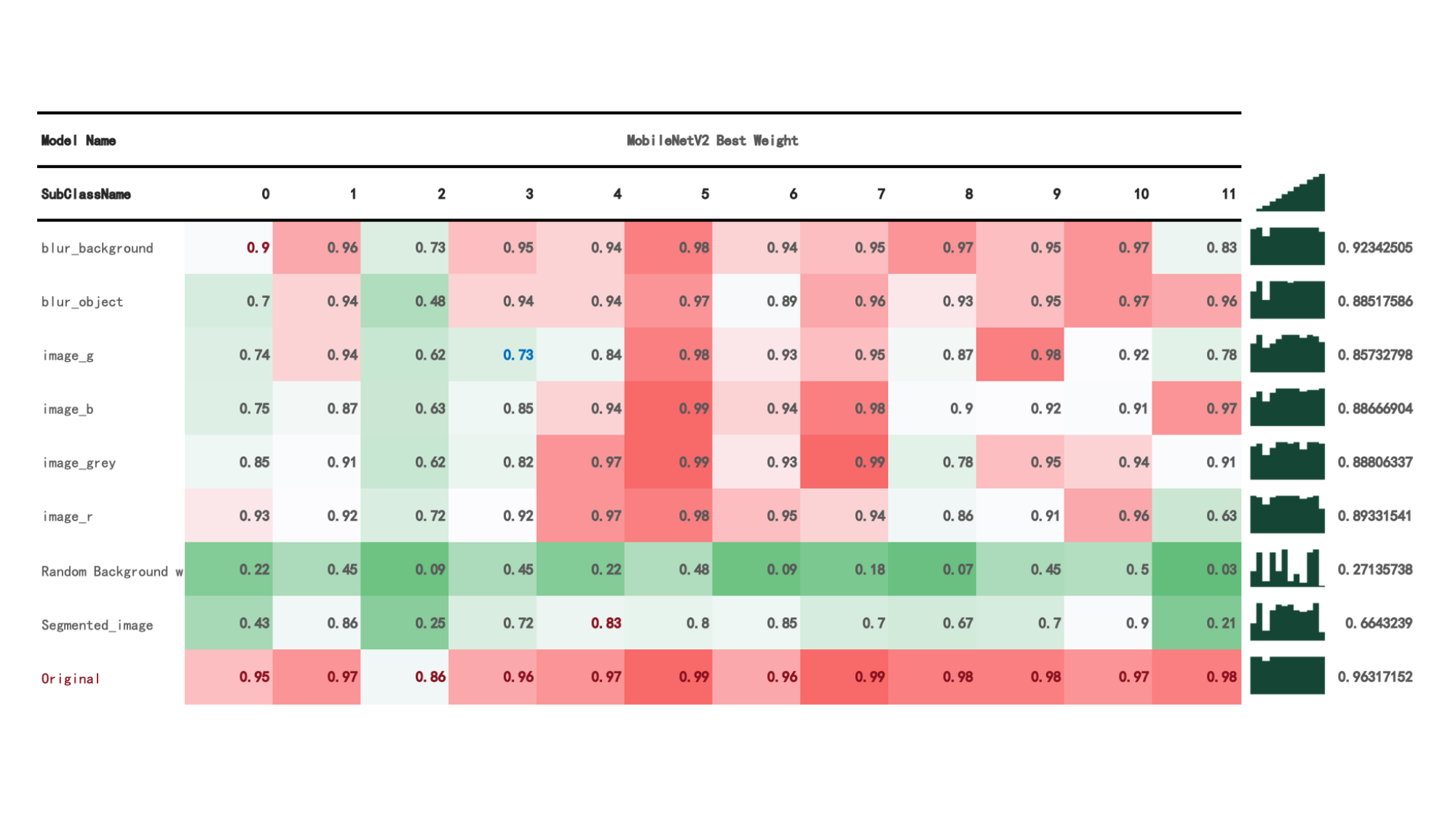}
        The MobileNetV2 \cite{sandler2018mobilenetv2} accuracy for each class
    \end{minipage}
    \hfill
    \begin{minipage}{0.3\textwidth}
        \centering
        \includegraphics[width=\linewidth]{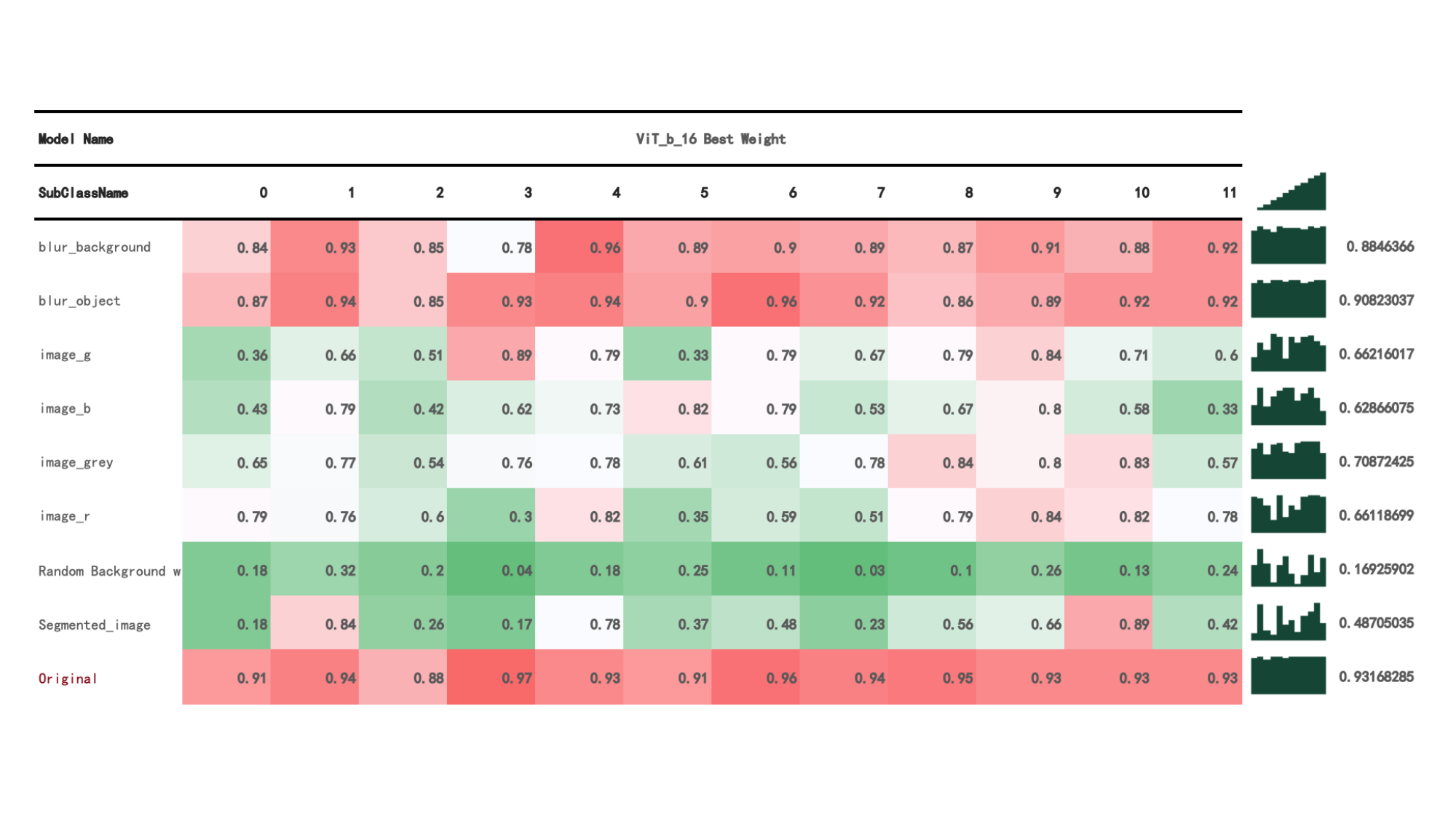}
        The ViT \cite{dosovitskiy2020image} accuracy for each class
    \end{minipage}
    \hfill
    \begin{minipage}{0.3\textwidth}
        \centering
        \includegraphics[width=\linewidth]{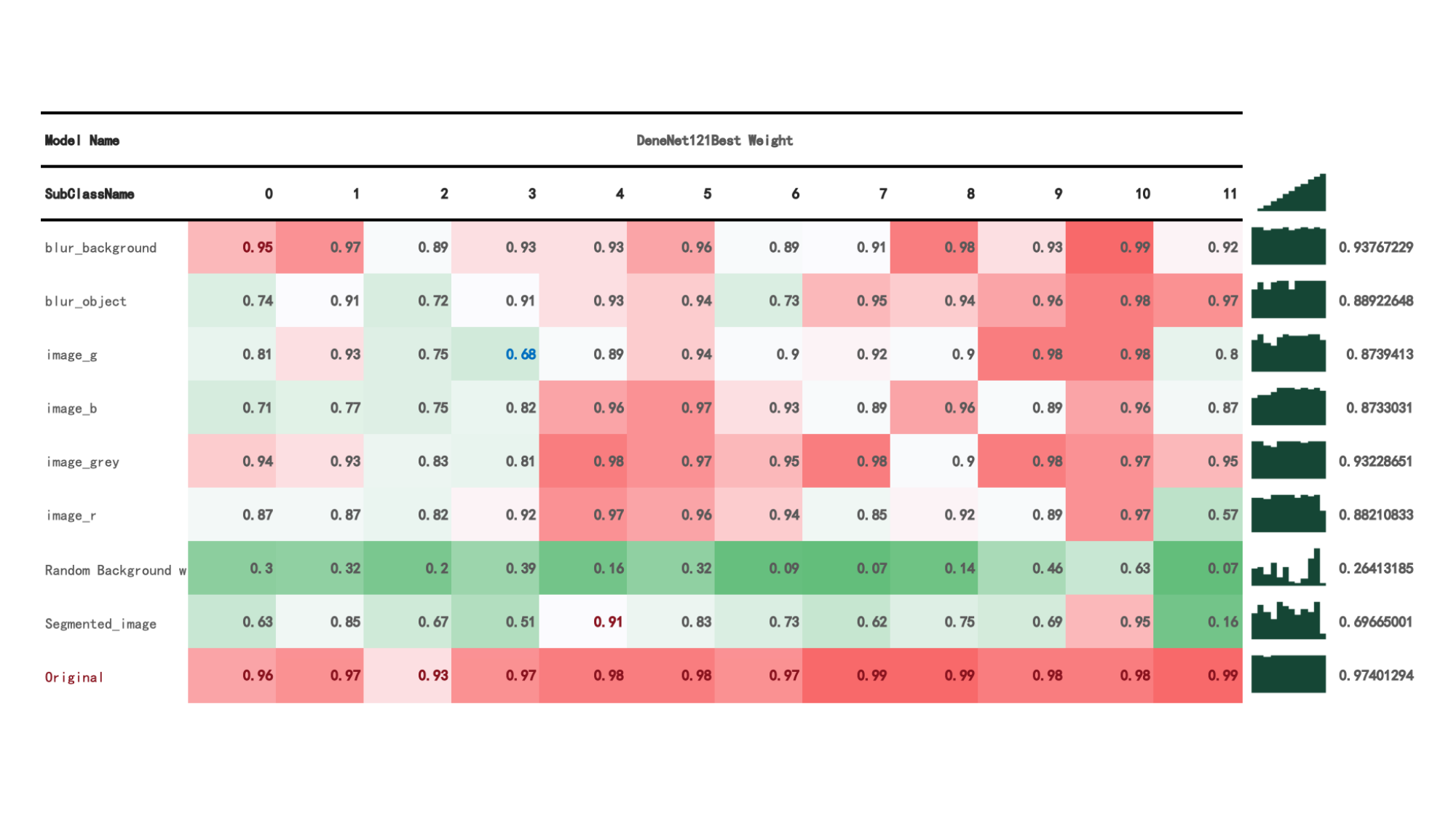}
        The DenseNet121 \cite{huang2017densely} accuracy for each class
    \end{minipage}
    \hfill
    \begin{minipage}{0.3\textwidth}
        \centering
        \includegraphics[width=\linewidth]{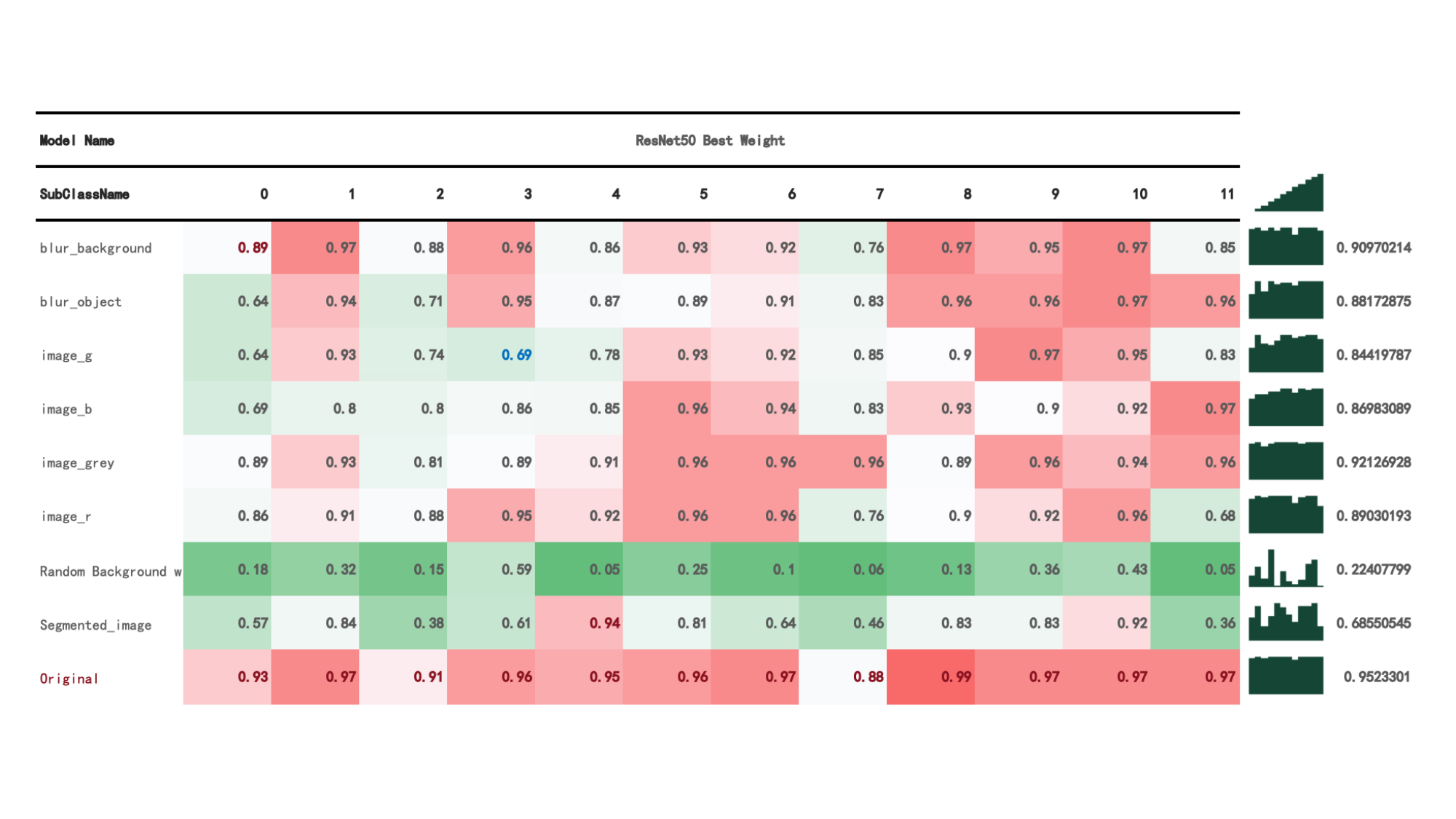}
        The ResNet50 \cite{he2016deep} accuracy for each class
    \end{minipage}
    \caption{The SOTA models accuracy density map for each class on Experiment 1. Testing images with different background indeed is a challenging scenario for vision models.}
    \label{densitymap}
\end{figure*}

\begin{figure*}[t]
    \centering
    \begin{minipage}{0.3\textwidth}
        \centering
        \includegraphics[width=\linewidth]{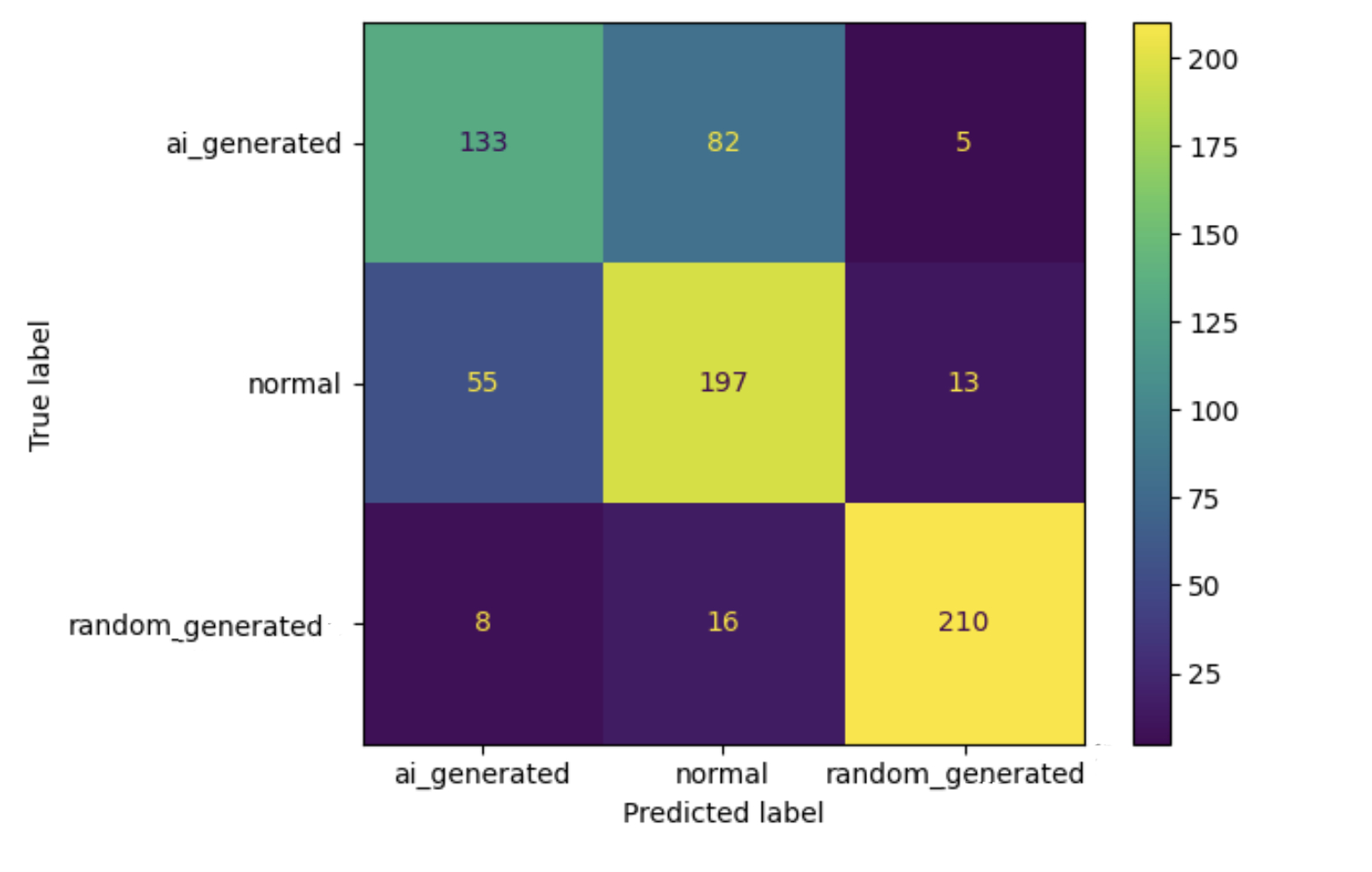}
        ResNet50 \cite{he2016deep} 74.9\%
    \end{minipage}
    \hfill
    \begin{minipage}{0.3\textwidth}
        \centering
        \includegraphics[width=\linewidth]{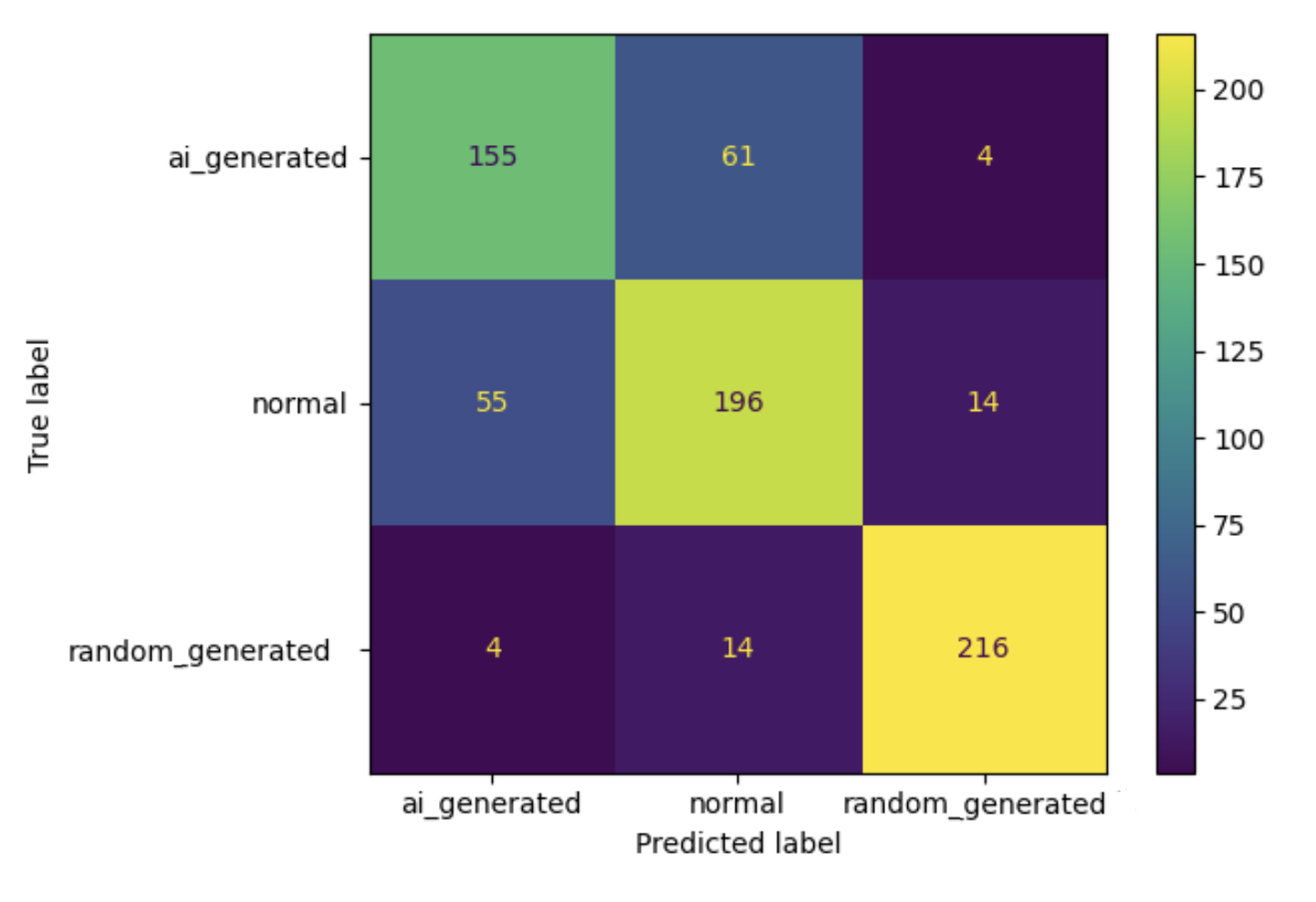}
        DenseNet \cite{huang2017densely} 78.8\%
    \end{minipage}
    \hfill
    \begin{minipage}{0.3\textwidth}
        \centering
        \includegraphics[width=\linewidth]{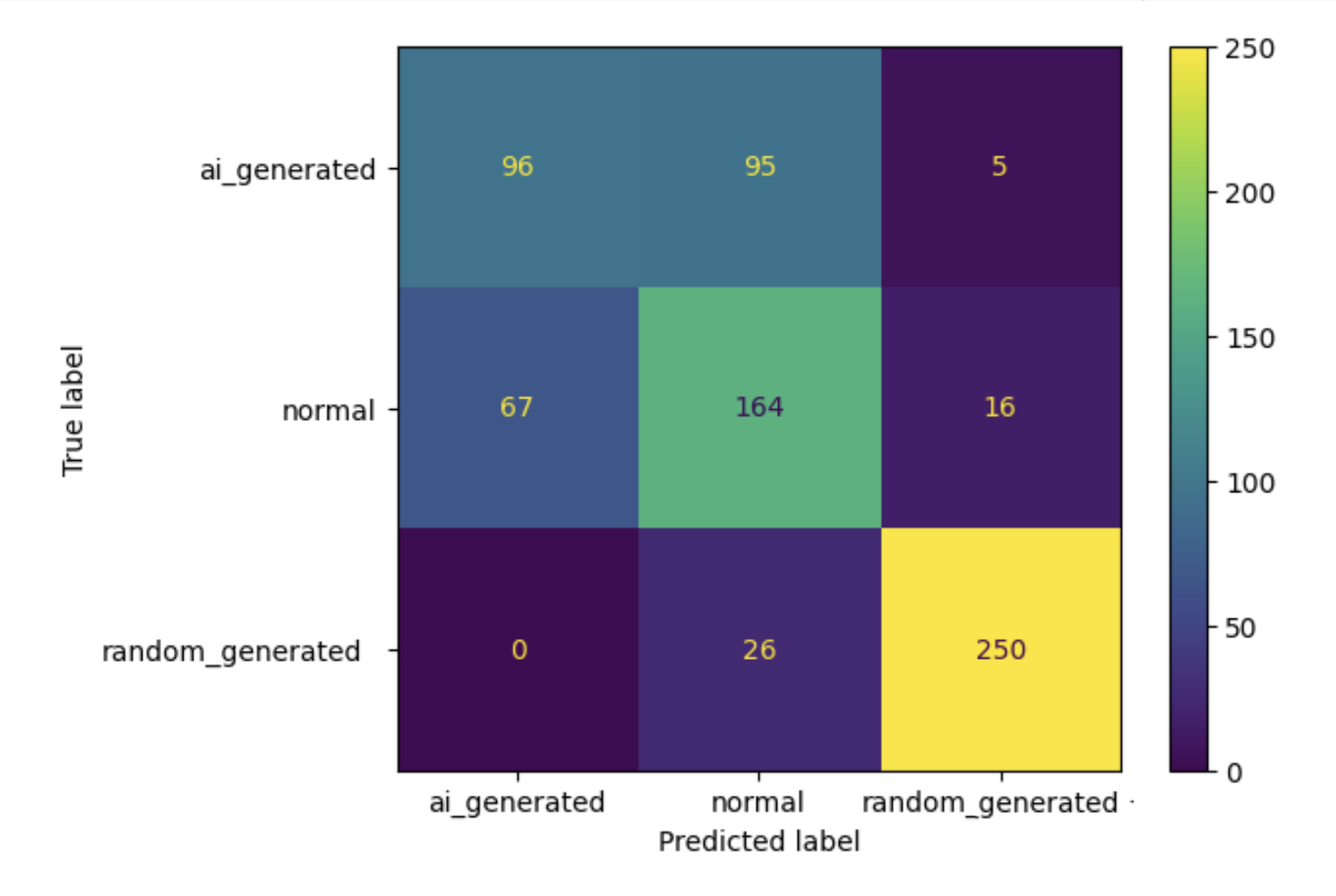}
        MobileNet \cite{sandler2018mobilenetv2} 70.9\%
    \end{minipage}
    \caption{Model accuracy for classifying normal/AI-generated/random-generated images.}
    \label{fig:type3-model-performance}  
\end{figure*}

\begin{figure*}[h]
    \centering
    \begin{minipage}{0.2\textwidth}
        \centering
        \includegraphics[width=1\linewidth]{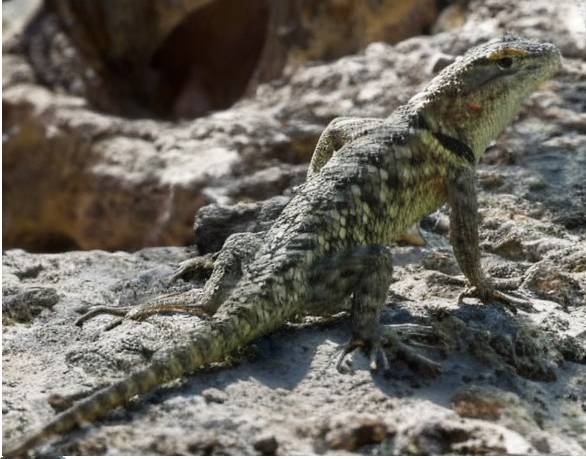}
        \textbf{Prompt}: Generate high-definition pictures like those in the National Geographic magazine, keep the background unchanged.
        \vspace{-8mm}
    \end{minipage}%
    \hfill
    \begin{minipage}{0.2\textwidth}
        \centering
        \includegraphics[width=1\linewidth]{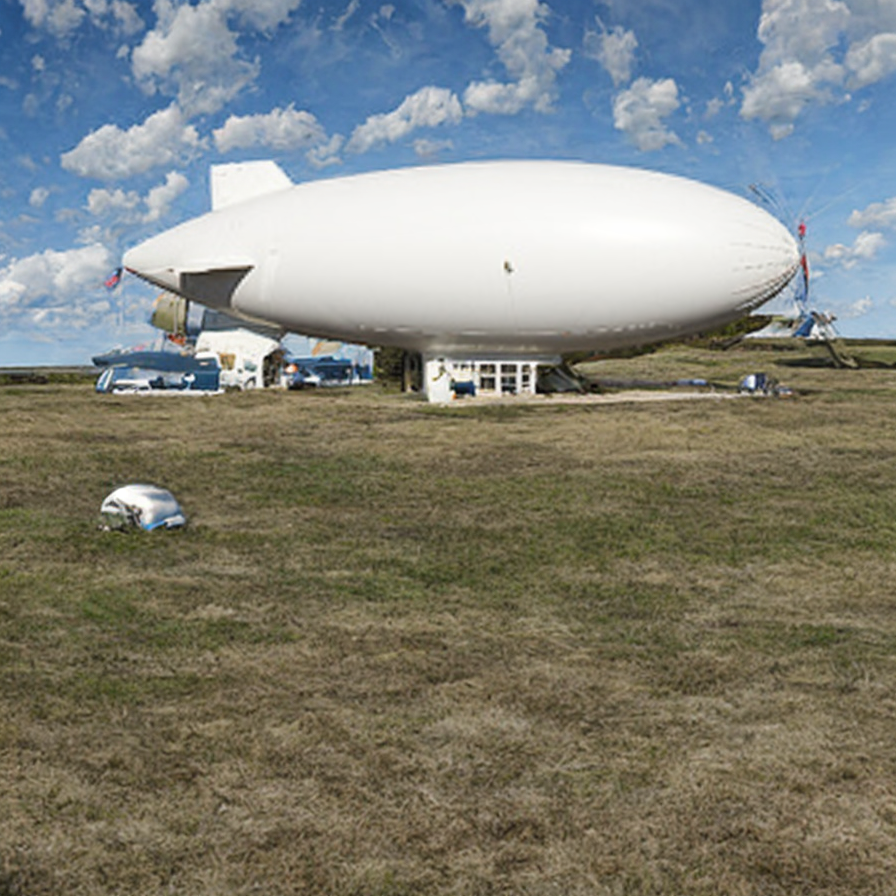}
        \textbf{Prompt}: Generate a realistic blue sky, and clouds background and please do not change the foreground airship object.
    \end{minipage}%
    \hfill
    \begin{minipage}{0.2\textwidth}
        \centering
        \includegraphics[width=1\linewidth]{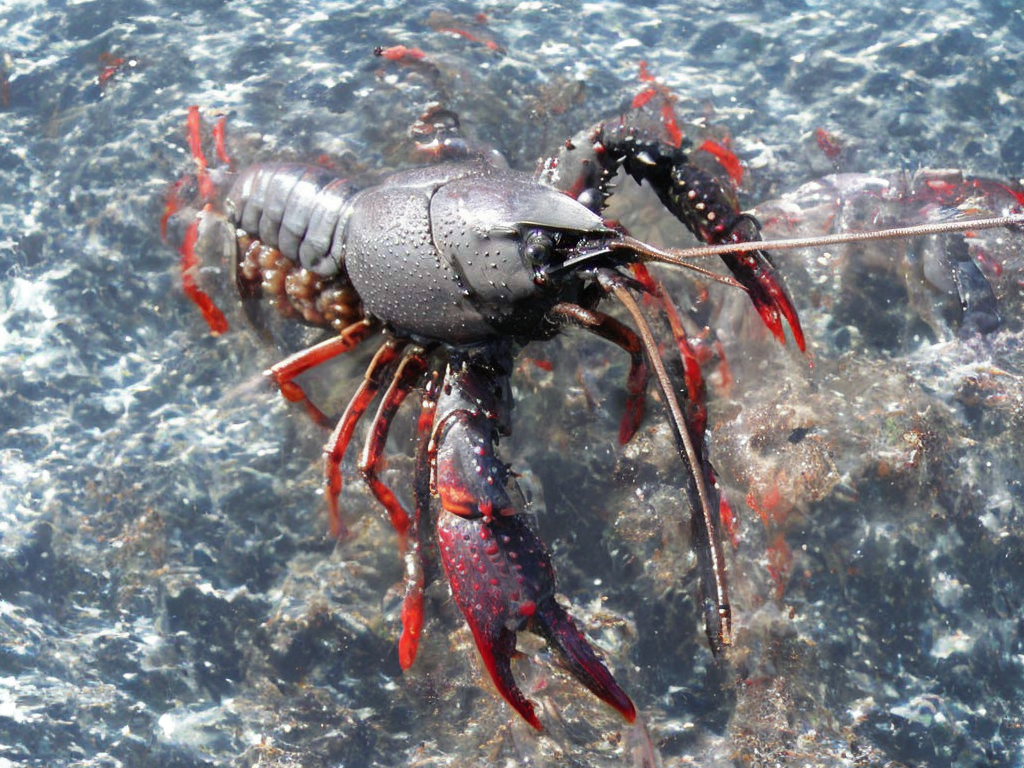}
        \textbf{Prompt}: Generate high resolution images in sea water.
        \vspace{4.5mm}
    \end{minipage}%
    \hfill
    \begin{minipage}{0.2\textwidth}
        \centering
        \includegraphics[width=1\linewidth]{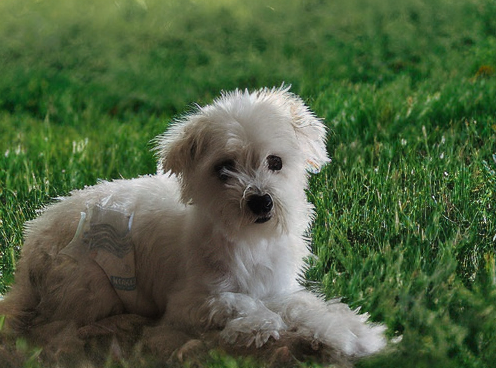}
        \textbf{Prompt}: Generate a picture with a foreground and the green grass in the background, similar to the official HD picture released by the state.
        \vspace{-13.2mm}
    \end{minipage}%
    \vspace{5mm}
    \begin{minipage}{0.2\textwidth}
        \centering
        \includegraphics[width=1\linewidth]{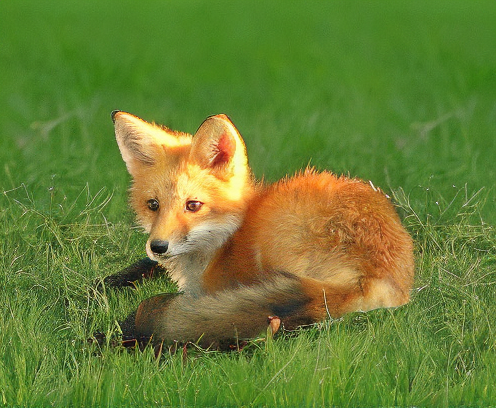}
        \textbf{Prompt}: Generate high-resolution pictures like fox in the lawn, National Geographic,  keep the background and foreground more simple and real.
        \vspace{5mm}
    \end{minipage}%
    \hfill
    \begin{minipage}{0.2\textwidth}
        \centering
        \includegraphics[width=1\linewidth]{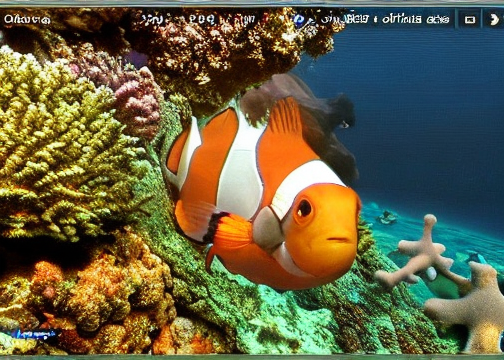}
        \textbf{Prompt}: Generate a simple image more realistic in the style of ocean magazine.
        \vspace{13.5mm}
    \end{minipage}%
    \hfill
    \begin{minipage}{0.2\textwidth}
         \centering
         \includegraphics[width=1\linewidth]{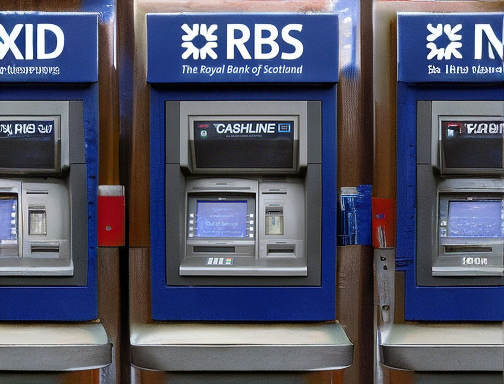}
         \textbf{Prompt}: Generate high-resolution pictures like those in National Financial Magazine, and keep the background and foreground consistent and the environment more real!
         \vspace{-2mm}
     \end{minipage}%
    \hfill
	 \begin{minipage}{0.2\textwidth}
       \centering
       \includegraphics[width=1\linewidth]{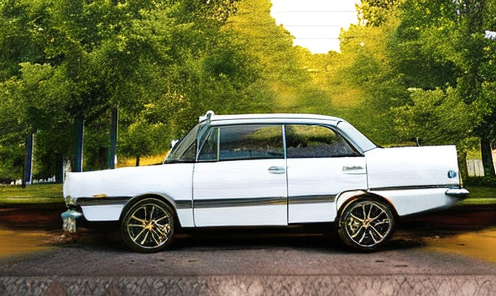}
       \textbf{Prompt}: Generate an image with the car in the background and similar to the HD image published by the state.
       \vspace{4mm}
   \end{minipage}%
   
	\begin{minipage}{0.2\textwidth}
	    \centering
	    \includegraphics[width=1\linewidth]{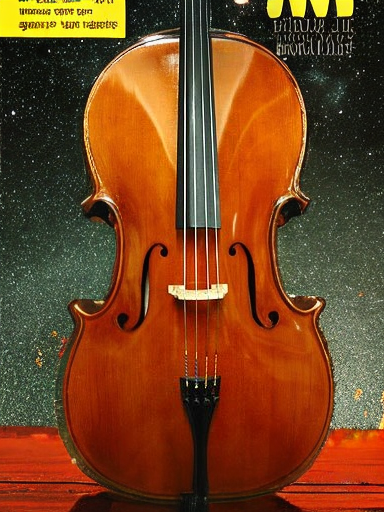}
	    \textbf{Prompt}: Please generate high-resolution pictures like those in the National Music Magazine and keep the background unchanged.
    \end{minipage}%
    \hfill
     \begin{minipage}{0.2\textwidth}
          \centering
          \includegraphics[width=1\linewidth]{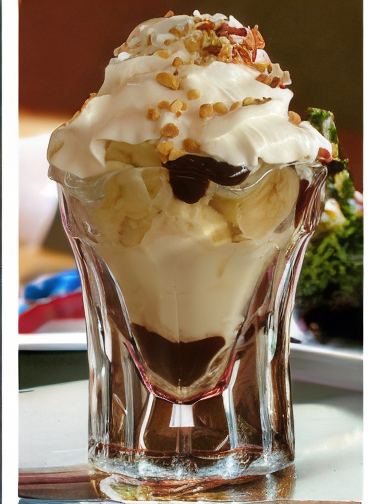}
          \textbf{Prompt}: Generate high-resolution pictures in the style of those in National Food Magazine, and keep the background and foreground consistent and the environment more real!
          \vspace{-12mm}
   	 \end{minipage}%
    \hfill
 	\begin{minipage}{0.2\textwidth}
        \centering
        \includegraphics[width=1\linewidth]{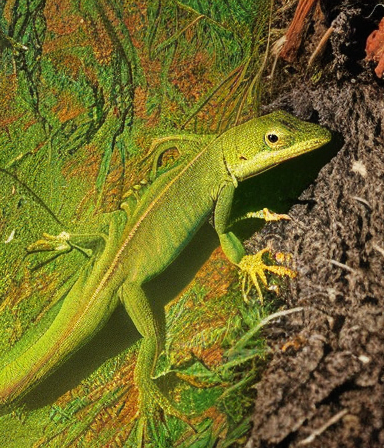}
        \textbf{Prompt}: Generate high-definition pictures like those in the National Geographic magazine, keep the background unchanged.
        \vspace{-7mm}
    \end{minipage}%
    \hfill
	\begin{minipage}{0.2\textwidth}
        \centering
        \includegraphics[width=1\linewidth]{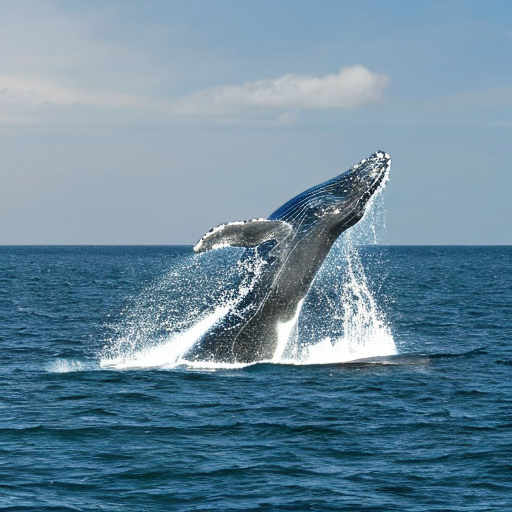}
        \textbf{Prompt}: Generate high-resolution pictures, such as National Marine Magazine's oceans and whales, to keep the background real.
        \vspace{-13mm}
    \end{minipage}%
    \vspace{10mm}
    \caption{AI generated images with prompts within XIMAGENET-12 Dataset.}
    \label{fig: prompt}
\end{figure*}

\end{document}